\documentclass{article}

\usepackage[preprint]{neurips_2026}

\usepackage[utf8]{inputenc} % allow utf-8 input
\usepackage[T1]{fontenc}    % use 8-bit T1 fonts
\usepackage{hyperref}       % hyperlinks
\usepackage{url}            % simple URL typesetting
\usepackage{booktabs}       % professional-quality tables
\usepackage{amsfonts}       % blackboard math symbols
\usepackage{nicefrac}       % compact symbols for 1/2, etc.
\usepackage{microtype}      % microtypography
\usepackage{xcolor}         % colors
\usepackage{amssymb}
\usepackage{graphicx}
\usepackage{subcaption}
\usepackage{multirow}
\usepackage[most]{tcolorbox}
\usepackage{amsmath}
\usepackage[table]{xcolor}  % 必须带table选项以支持表格行着色
\usepackage{pifont}

\def\eg{\emph{e.g.}}

\def\name{VAE-LFA}

\title{Why Do DiT Editors Drift? Plug-and-Play Low Frequency Alignment in VAE Latent Space}

\newcommand{\inst}[1]{$^{#1}$}
\newcommand{\spa}{\inst{\spadesuit}}
\newcommand{\hea}{\inst{\heartsuit}}
\newcommand{\dia}{\inst{\diamondsuit}}
\newcommand{\clu}{\inst{\clubsuit}}
\newcommand{\tri}{\inst{\triangle}}
\newcommand{\sta}{\inst{\star}}

\definecolor{ReferenceGray}{HTML}{F7F7F3}
\definecolor{QuantLavender}{HTML}{F2EFF9}
\definecolor{PruneGreen}{HTML}{EEF6F0}
\definecolor{JointBlue}{HTML}{E9F2FA}
\definecolor{RelAccBlue}{HTML}{315C99}

\newcommand{\prow}{\rowcolor{PruneGreen}}

\definecolor{darkgreen}{RGB}{0,100,0}

\author{%
  Xiaoce Wang$^{\dag}$ \spa \quad Sifan Zhou$^{\dag}$ \hea \quad Kaifei Wang$^{\dag}$ \dia \quad Leli Xu \spa \quad \\ \textbf{Xuerui Qiu} \clu \quad \textbf{Tao He} \tri \quad \textbf{Ming Li} \sta \\
  \spa Tsinghua University \quad \hea Carnegie Mellon University \quad \dia Peking University \\ 
  \clu CASIA \quad \tri University of Electronic Science and Technology of China \\ \sta Guangming Laboratory\\
  $^{\dag}$These authors contributed equally. \\
  \texttt{wangxc23@mails.tsinghua.edu.cn} \\
}

\begin{document}

\maketitle

\begin{center}
  \vspace{-3mm}
  \includegraphics[width=0.85\textwidth]{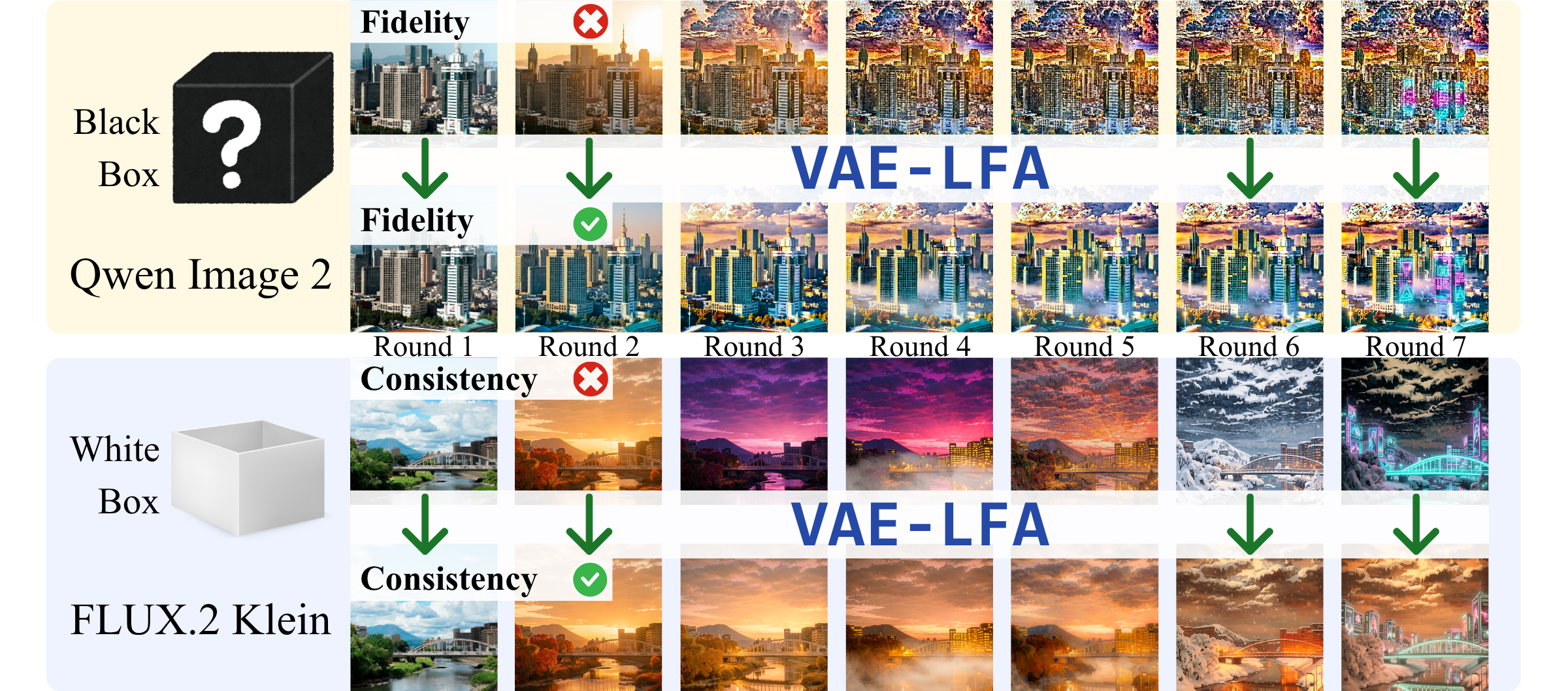}

  \begin{minipage}{0.8\textwidth}
    \captionof{figure}{\textbf{VAE-LFA improves multi-turn editing capabilities.}
    Black-box: background visual quality degrades severely; White-box: style consistency is suboptimal. VAE-LFA improves both.}
    \label{fig:teaser}
  \end{minipage}
  \vspace{-3mm}
\end{center}

\begin{abstract}
Recent advances in diffusion transformers (DiTs) have enabled promising single-turn image editing capabilities. However, multi-turn editing often leads to progressive semantic drift and quality degradation.
In this work, we study this problem from a latent-space frequency perspective by decomposing the editing process into two functional components: VAE and DiT. Through systematic analysis in the VAE latent space, we uncover that the DiT introduces dominant low-frequency drift that accumulates as semantic misalignment across editing rounds, while the VAE contributes comparatively stable reconstruction bias.
Based on this insight, we propose \textbf{VAE-LFA} (Low Frequency Alignment), a training-free, plug-and-play method that performs alignment in VAE latent space. VAE-LFA decomposes latent discrepancies across editing rounds via low-pass filtering, and aligns low-frequency statistics to an exponential moving average of previous rounds, effectively suppressing accumulated semantic drift while preserving high-frequency details.
Our method requires \textit{no retraining, ground-truth priors, or access to diffusion parameters}, making it applicable to both white-box and black-box DiT editors. For white-box models, VAE-LFA is seamlessly integrated into the editing pipeline by eliminating redundant VAE round trips; for black-box models, it operates via an off-the-shelf VAE to perform inter-round latent alignment.
Extensive experiments demonstrate that VAE-LFA improves semantic consistency and visual fidelity across diverse multi-turn editing scenarios, including both controlled and in-the-wild images.
  
\end{abstract}

\vspace{-0.1cm}
\section{Introduction}
\vspace{-0.2cm}
Recent advances in instruction-based image editing models built upon diffusion transformers (DiTs) and flow matching have achieved remarkable improvements in prompt-following, realism, and visual quality for single-round editing~\cite{rombach2022high, labs2025flux, wu2025qwen, brooks2023instructpix2pix, geng2024instructdiffusion, huang2024smartedit, fu2023guiding, zhang2023magicbrush}. Despite these advances, their performance degrades significantly under multi-turn (\eg, 5+ turns) iterative editing, often leading to severe semantic drift and quality deterioration \cite{zhou2025multi, liao2025freqedit, kim2025improving}. This limitation poses a critical challenge for real-world deployment as users cannot be expected to articulate all editing intents with sufficient precision in a single prompt, and iterative refinement is inherent to any creative workflow.
While prior works have proposed various remedies for multi-turn degradation, they either demand full-model retraining~\cite{qu2025vincie, sheynin2024emu, almog2025reed}, require internal access to denoising trajectories~\cite{kim2025improving}, or rely on model-specific inference signals such as internal features or velocity fields~\cite{liao2025freqedit}. Nevertheless, the community heavily relies on black-box editing models like GPT Image, Qwen Image 2.0~\cite{wu2025qwen}, Seedream 4.0~\cite{seedream2025seedream} and GLM-Image~\cite{glmimage}, yet these methods cannot be directly deployed on black-box or API-based editors without modification. This calls for a simple and effective remedy for editing degradation that \textbf{requires no internal access.}

% To rigorously attribute the source of drift, we argue that the \textbf{VAE latent space} offers a principled vantage point. A VAE encoder-decoder bottleneck is shared by virtually all modern diffusion-based editors, making it a universal inspection point. Moreover, recent work has established a consistent correspondence between frequency bands in the latent space and their semantic counterparts in image space~\cite{ning2026spectrummatchingunifiedperspective, medi2025missingfinedetailsimages}, enabling interpretable frequency decomposition without requiring access to complicated internal DiT parameters.

% Moreover, fundamental questions remain unanswered: \textit{Why do DiT editors drift? Which component is the primary cause?}
% %
% To address this question, we argue that the \textbf{VAE latent space} provides a principled and model-agnostic vantage point. The VAE encoder-decoder bottleneck is shared across virtually all modern diffusion-based editors, making it a natural and universal interface for analysis. Moreover, recent studies have established a consistent correspondence between frequency bands in latent space and their semantic counterparts in image space~\cite{ning2026spectrummatchingunifiedperspective, medi2025missingfinedetailsimages}, enabling interpretable frequency-based analysis without requiring access to internal DiT representations.

\begin{figure*}[h]
  \centering
  \includegraphics[width=0.8\linewidth]{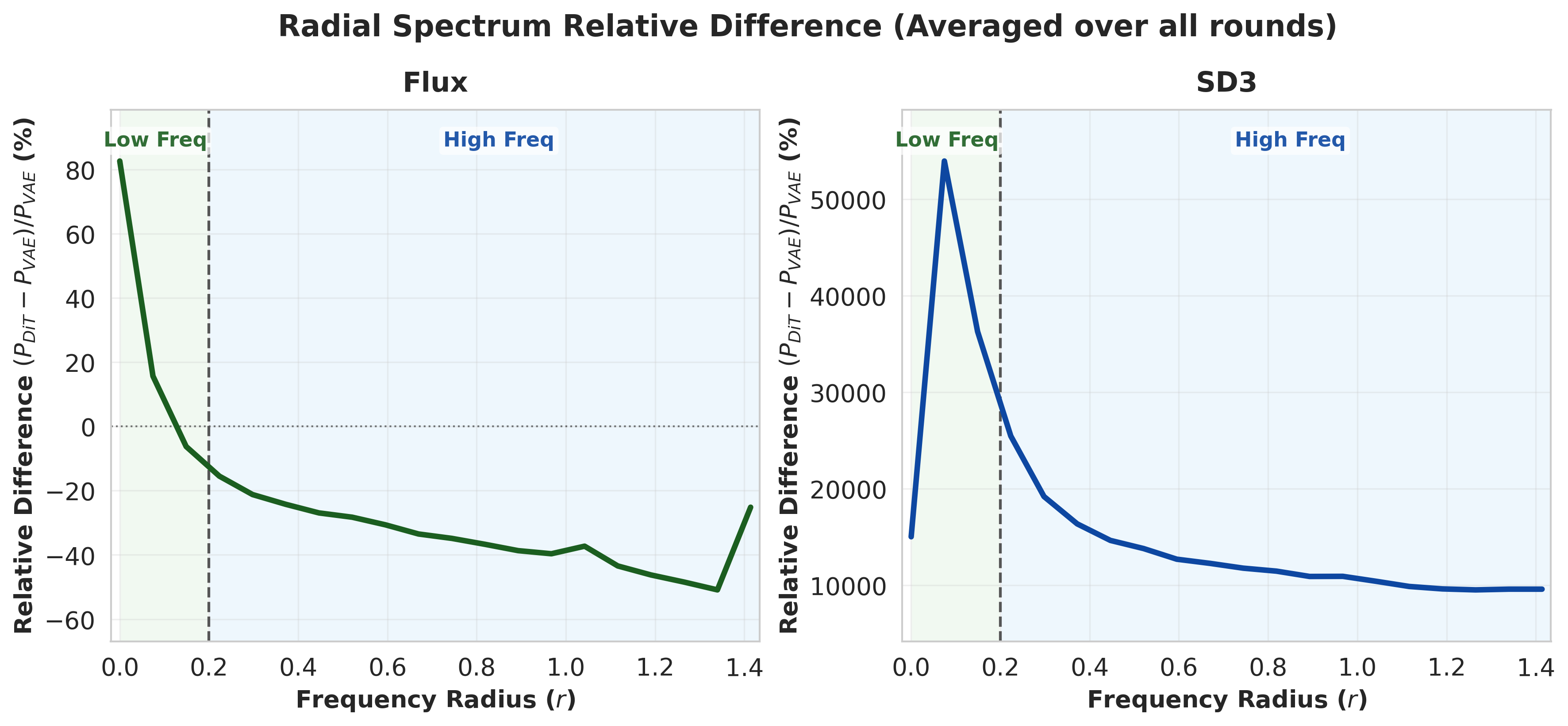}
  \vspace{-1mm}
  \caption{\textbf{Cross-architecture validation of frequency-wise drift.}
  Relative radial-spectrum difference between DiT-only and VAE-only loops over 10 rounds:
  $\Delta P(r)=\frac{P_{\text{DiT}}(r)-P_{\text{VAE}}(r)}{P_{\text{VAE}}(r)}\times100\%$.
  On both FLUX.2 and SD3, DiT transitions dominate low-frequency drift ($r<0.2$), while VAE bias is stronger at higher frequencies. The vertical line marks $r=0.2$.}
  \label{fig:radial_spectrum}
  \vspace{-2mm}
\end{figure*}

Moreover, fundamental questions remain unanswered: \textit{Why do DiT editors drift? Which component is the primary cause?}
To answer these questions, we draw upon recent evidence that frequency-based analysis in latent space provides a principled perspective~\cite{ning2026spectrummatchingunifiedperspective, medi2025missingfinedetailsimages}.
We focus on the \textbf{VAE latent space}, as the VAE encoder--decoder bottleneck is a fundamental component shared across virtually all modern DiT-based editors, making it a universal interface for component-level drift attribution.

Specifically, we decompose the editing pipeline into two components, the VAE for encoding and decoding, and the DiT for iterative refinement, and systematically analyze frequency-wise drift in the latent space using no-operation prompts to isolate intrinsic degradation. As shown in Fig.~\ref{fig:radial_spectrum} and Fig.~\ref{fig:vae_reduction}, our analysis reveals a striking asymmetry: semantic degradation is predominantly driven by accumulated low-frequency deviations introduced by the DiT over multiple iterations, whereas the VAE loop merely induces a stable, spectrally uniform reconstruction bias. This finding is counter-intuitive, challenging the prevailing hypothesis that repeated VAE cycles dominate multi-turn error accumulation~\cite{almog2025reed, yu2026i2eimagepixelsactionable}. We further discover that VAE acts as an implicit regularizer that partially curbs DiT-induced low-frequency drift, rather than purely introducing degradation, as shown in Fig.~\ref{fig:vae_reduction}.

\begin{figure}
  \centering
  \includegraphics[width=\linewidth]{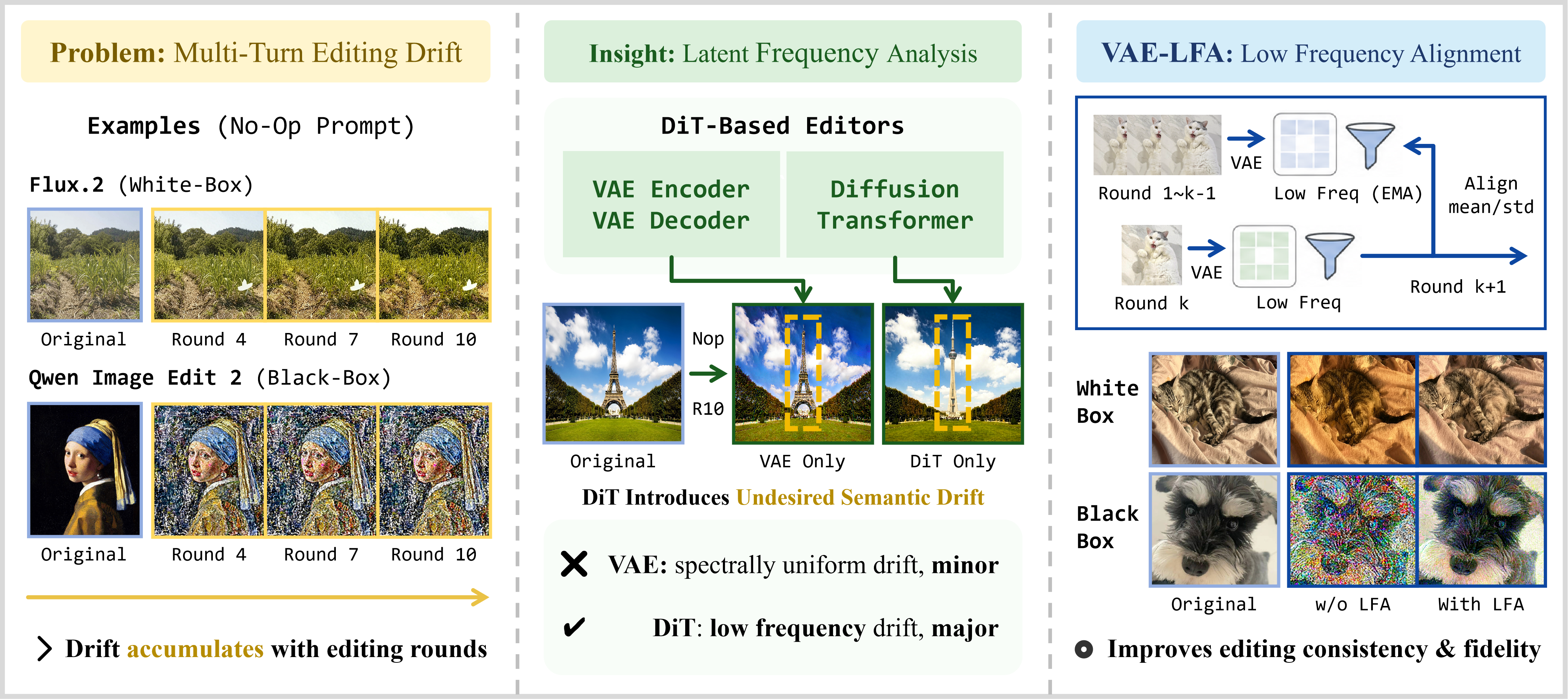}
  \caption{\textbf{Overview of VAE-LFA}: component-wise drift analysis and low-frequency latent alignment.}
  \label{fig:main}
  \vspace{-4mm}
\end{figure}

Motivated by this asymmetry, we propose \textbf{VAE-LFA} (Low Frequency Alignment) (Fig.~\ref{fig:main}), a training-free and plug-and-play framework that operates directly in the VAE latent space. The key idea is to selectively constrain the low-frequency components that dominate semantic drift, while leaving high-frequency components unconstrained to preserve fine details and editing flexibility. By stabilizing the low-frequency latent statistics across editing rounds, our method effectively mitigates the accumulation of semantic misalignment without restricting the model’s ability to perform meaningful edits.
Crucially, \name{} requires no retraining, no ground-truth priors, and no access to internal model parameters, making it applicable to both white-box and black-box DiT-based editors. This design enables seamless integration into existing editing pipelines, either by directly operating within the model (for white-box settings) or by interfacing through an external latent space (for black-box settings), without modifying the underlying model. Extensive experiments demonstrate that \name{} mitigates quality degradation across diverse editing scenarios. Across no-op, cycle, and long-chain protocols, our method improves semantic consistency and perceptual fidelity while preserving instruction-following capability (Fig.~\ref{fig:teaser}, Fig.~\ref{fig:rounds_curve}, Tab.~\ref{tab:noop_cycle}, and Tab.~\ref{tab:longchain}). Overall, our contribution can be summarized as:
\begin{itemize}
    \item We present an interpretable analysis of multi-turn editing degradation from a VAE latent frequency perspective, showing that semantic drift is mainly driven by low-frequency drift.
    \item We provide a controlled decomposition of the editing pipeline, revealing an asymmetry between DiT and VAE, where the former dominates semantic drift while the latter introduces stable reconstruction bias with partial regularization effects.
    \item We propose \name{}, a training-free and plug-and-play method that mitigates drift via low-frequency latent alignment, requiring no retraining or internal model access and supporting both white-box and black-box DiT-based editors with consistent performance improvements.
\end{itemize}

\vspace{-3mm}
\section{Related Works}
\vspace{-3mm}

Instruction-based DiT editors achieve strong single-round results~\cite{rombach2022high, labs2025flux, wu2025qwen}, yet suffer from progressive drift under multi-turn editing~\cite{zhou2025multi, liao2025freqedit, kim2025improving}. Prior remedies require retraining~\cite{qu2025vincie, sheynin2024emu}, internal access~\cite{kim2025improving}, or specialized architectures~\cite{liao2025freqedit}, excluding commercial black-box models~\cite{wu2025qwen, seedream2025seedream, glmimage}. While the VAE bottleneck is known to introduce reconstruction bias~\cite{yu2026i2eimagepixelsactionable, kwon2025vision}, existing understanding of its latent space remains largely theoretical~\cite{rombach2022high, kouzelis2025eqvaeequivarianceregularizedlatent, gu2026understandinglatentdiffusabilityfisher}, with limited insight into how VAE latent frequency bands govern specific semantics. Recent frequency-semantic correspondences~\cite{ning2026spectrummatchingunifiedperspective, medi2025missingfinedetailsimages} offer a promising lens, but have not been exploited to attribute or mitigate iterative drift.

In this work, we aim to \textbf{explain and solve} multi-turn editing degradation through a principled, interpretable analysis of the VAE latent space. By decomposing drift into frequency-specific components, we reveal which pipeline stage corrupts which semantic band, and why. Building on this causal understanding, we develop a training-free remedy that operates without internal model access, making it applicable to both open-source and black-box DiT-based editors. A detailed discussion of related works is provided in Appendix~\ref{appendix:related_works}.

\vspace{-0.1cm}
\section{Observations}
\label{sec:observations}
\vspace{-0.2cm}
% For brevity, we refer to the generative backbone of both diffusion and flow-matching editors as DiT hereafter, since they share the same transformer architecture and operate on similar VAE latent spaces.
% \vspace{-0.1cm}
\subsection{Formulation: Multi-turn Editing Bias}
\label{sec:formulation}
We define multi-turn editing bias as the fixed-point violation of an editor under no-operation prompts: ideally, $\mathcal T_{p_{\mathrm{no\text{-}op}}}(\mathbf{x})=\mathbf{x}$, while in practice repeated no-op editing accumulates semantic drift.

In the VAE latent space, let $E,D$ denote the VAE encoder/decoder and $G_p$ the prompt-conditioned DiT transition, with $\Phi_p(\mathbf z)=E(D(G_p(\mathbf z)))$. The no-op latent bias decomposes into
\begin{equation}
\Phi_{p_{\mathrm{no\text{-}op}}}(\mathbf z)-\mathbf z
=
\underbrace{G_{p_{\mathrm{no\text{-}op}}}(\mathbf z)-\mathbf z}_{\text{DiT bias}}
+
\underbrace{\Phi_{p_{\mathrm{no\text{-}op}}}(\mathbf z)-G_{p_{\mathrm{no\text{-}op}}}(\mathbf z)}_{\text{VAE round-trip bias}} .
\end{equation}
This motivates our component-wise analysis, since small per-turn biases can accumulate over long trajectories. A detailed formulation of prompts and multi-turn editing bias is provided in Appendix~\ref{appendix:formulation}.

\subsection{Viewing Bias from a Frequency Perspective}
\vspace{-0.1cm}
\label{sec:frequency_perspective}
\paragraph{Frequency decomposition in latent space.}
A VAE latent $\mathbf{z}\in\mathbb R^{C\times H\times W}$ can be viewed as $C$ two-dimensional spatial feature maps. We define frequency by applying a 2D Fourier transform to each channel independently over the spatial dimensions $(H,W)$, while the channel dimension is not transformed. Under this view, each latent channel contains a zero-frequency component and non-zero-frequency components, also known as DC and AC components, respectively.

For a latent tensor $\mathbf{u}\in\mathbb R^{C\times H\times W}$, we compute channel-wise spatial statistics
\begin{equation}
\begin{aligned}
    \mu_c(\mathbf{u})=\frac{1}{HW}\sum_{i,j}\mathbf{u}_{c,i,j},\quad\sigma_c(\mathbf{u})=\bigg(\frac{1}{HW}\sum_{i,j}\big(\mathbf{u}_{c,i,j}-\mu_c(\mathbf{u})\big)^2\bigg)^{1/2}.
\end{aligned}
\label{eq:latent_mean_std}
\end{equation}
The mean $\mu_c(\mathbf{u})$ corresponds to the DC component of channel $c$ up to the Fourier normalization factor. After removing this DC component, the centered signal $\mathbf{u}_c-\mu_c(\mathbf{u})$ contains only AC components. By Parseval's identity, $\sigma_c^2(\mathbf{u})$ is proportional to the total energy of the non-DC Fourier coefficients, and thus measures the average AC energy of channel $c$.

We then define low- and high-frequency latent components using a spatial low-pass filter. Let $L$ denote a low-pass filter applied independently to each latent channel. A latent $\mathbf{z}$ is decomposed as
\begin{equation}
    \mathbf{z}_{\mathrm{low}}=L(\mathbf{z}),
    \quad
    \mathbf{z}_{\mathrm{high}}=\mathbf{z}-L(\mathbf{z}).
    \label{eq:low_high_latent}
\end{equation}

% Decode -> Approximately Additive -> One-to-one solution
\subsection{Empirical Analysis of Low and High Frequency Components}
\vspace{-0.1cm}

\begin{figure*}[t]
  \centering
  \includegraphics[width=\linewidth]{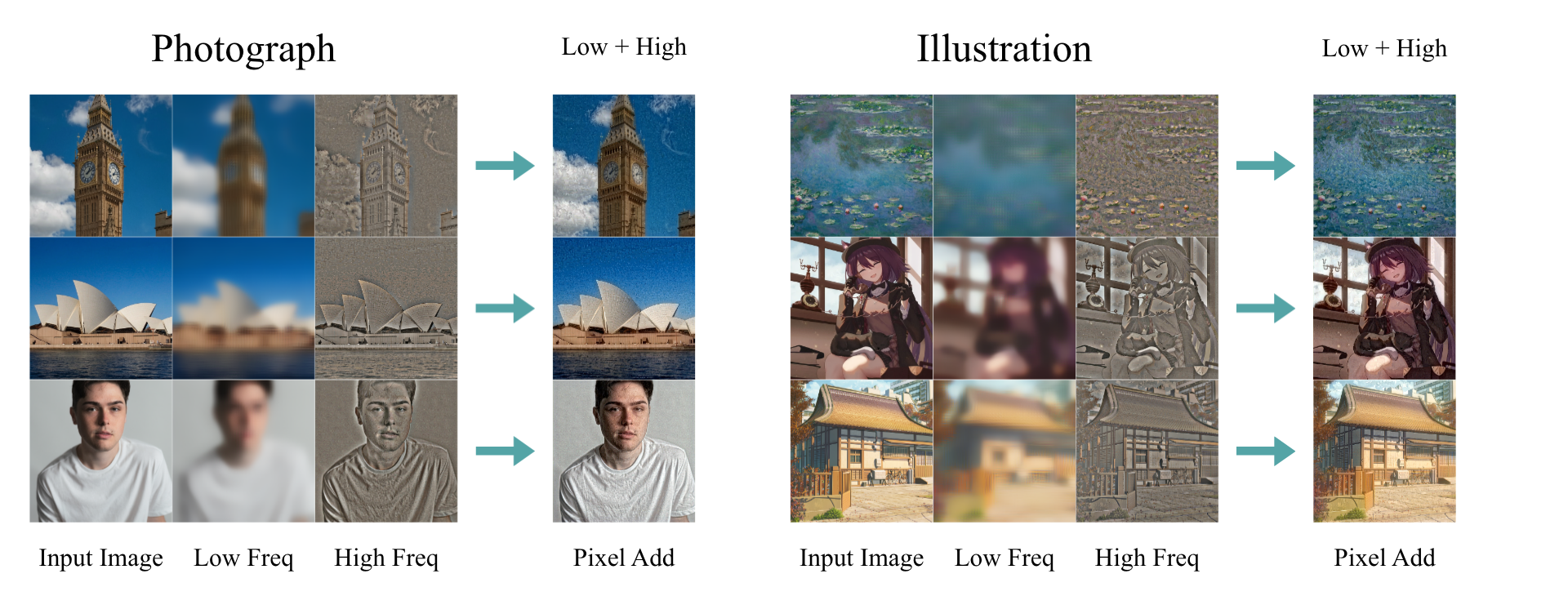}
  \caption{\textbf{Low/high-frequency components in the FLUX.2 VAE latent space.} Low frequencies capture global appearance (e.g., color); high frequencies retain local details (e.g., texture).}
  \label{fig:freq_vis}
  \vspace{-4mm}
\end{figure*}

To analyze the specific characteristics of low and high frequency components, we employ a filter in the VAE latent space of FLUX.2~\cite{flux-2-2025}, a DiT-based editor, to isolate the two, before decoding them back into pixel space. As demonstrated in Fig.~\ref{fig:freq_vis}, we identify that the low-frequency clearly represents \textbf{macro information}, such as tone, style and overall structure, while high frequency components represent \textbf{micro details}, such as textures and outlines. 

While VAE decoders are nonlinear maps from latent space to image space~\cite{kingma2013auto}, our experiments in Fig.~\ref{fig:freq_vis} suggest approximate additivity for the frequency decomposition considered here. Specifically, the visual effects observed by decoding low and high frequency components separately are consistent to a degree with their roles in the jointly decoded image. This makes separate low/high decoding a meaningful diagnostic of their image space representations. It also supports our choice to process the low and high frequency residuals separately, before recombining them by simple latent addition.

% Separate VAE/DiT -> Hi Freq VAE, low freq DiT -> VAE operation allev. DiT low freq
\subsection{Disentangling VAE and DiT in VAE Latent Space}
\vspace{-0.1cm}

\begin{figure*}[t]
  \centering
  \includegraphics[width=0.9\linewidth]{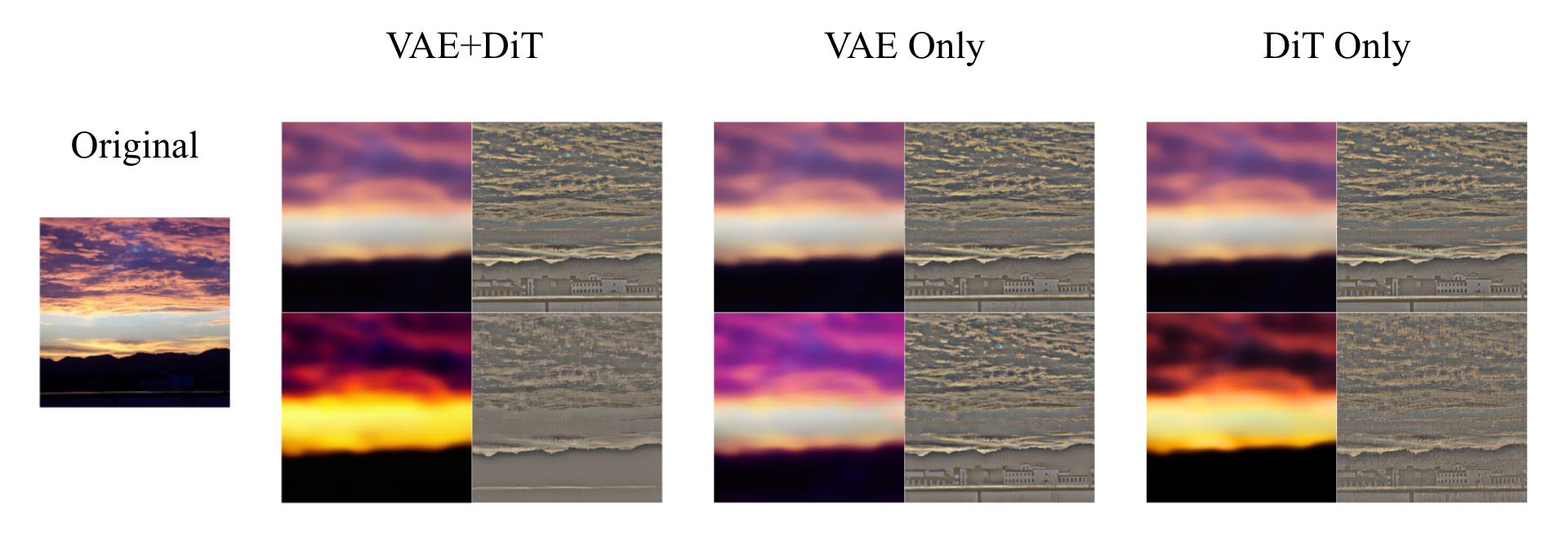}
  \vspace{-0.3cm}
  \caption{Low/high-frequency drift between round 1 (up) \& 10 (down) under three different settings.}
  \label{fig:freq_r10}
  \vspace{-0.3cm}
\end{figure*}

\paragraph{VAE-induced drift and DiT-induced drift.} We conduct spectral analysis of the FLUX.2 and Stable Diffusion 3 VAE latent space by implementing an average pooling filter to separate low and high frequency components. The experiment settings are:
\vspace{-0.1cm}
\begin{itemize}
\item \textbf{DiT only.} No-op editing passing from DiT to DiT without inter-round VAE decode-encodes.
\item \textbf{VAE only.} VAE encode-decode cycles, without the interference of DiT.
\end{itemize}
\vspace{-0.1cm}
We perform 10 rounds of experiments (Fig.~\ref{fig:radial_spectrum}). It is evident that DiT by itself contributes more significantly towards the low-frequency component. We then conduct visualizations of the low/high frequency status of the two settings, alongside the default baseline, as demonstrated in Fig.~\ref{fig:freq_r10}.
\vspace{-0.1cm}

\begin{figure*}[t]
  \centering
  \includegraphics[width=0.9\linewidth]{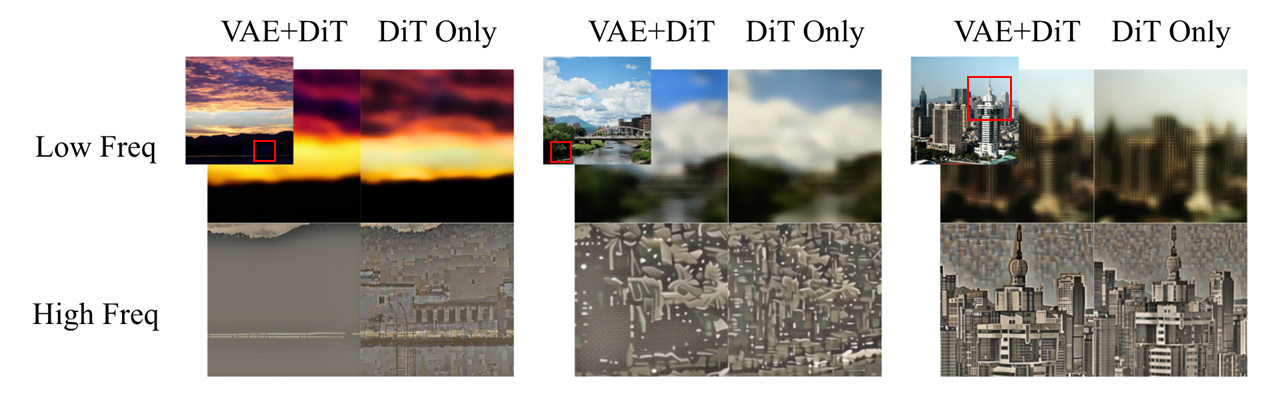}
  \vspace{-0.3cm}
  \caption{Removing VAE round trips improves high-frequency fidelity but leaves low-frequency drift.}
  \label{fig:vae_reduction}
  \vspace{-4mm}
\end{figure*}

\paragraph{Motivations of VAE-LFA.}
\vspace{-0.1cm}
As shown in Fig.~\ref{fig:vae_reduction}, removing intermediate VAE encode--decode operations substantially reduces high-frequency drift, but has a weaker effect on low-frequency drift. In fact, the inter-round VAE round trip sometimes preserves global attributes such as color more stably, suggesting that the VAE may partially regularize low-frequency drift.
Motivated by the VAE objective, which encourages the latent posterior to match a Gaussian prior parameterized by mean and variance~\cite{kingma2013auto}, we align the channel-wise mean and standard deviation of the low-frequency component. In the frequency domain, the mean corresponds to the DC component, i.e., the global channel-wise offset, while the variance controls the energy scale of centered non-DC components. Thus, mean/std alignment calibrates low-frequency statistics without directly constraining high-frequency residuals, leaving room for valid edits.
\vspace{-0.1cm}

\vspace{-2mm}
\section{Method}
\vspace{-2mm}
\subsection{VAE-LFA}
\label{sec:vae_lfa}
\vspace{-2mm}
Based on our observations, DiT is the primary source of semantic drift, while VAE introduces stable, spectrally uniform noise. This makes the VAE component easier to handle: when possible, we remove inter-round VAE encode-decode round trips. Therefore, VAE-LFA mainly alleviates the degradation induced by DiT by aligning the low-frequency components of the VAE latent space.

\paragraph{Mitigating drift induced by DiT.}
VAE-LFA inserts an alignment step in the VAE latent space after the DiT transition and before the next editing turn. Following the frequency decomposition in Sec.~\ref{sec:frequency_perspective}, we instantiate the low-pass operator $L$ as channel-wise average pooling: $L(\mathbf z)=\mathrm{AvgPool}_{\rho}(\mathbf z)$,
where $\rho$ is the pooling window size. For the $k$-th turn, we apply this filter to the DiT output latent and decompose it into a low-frequency component and a high-frequency residual as: $\boldsymbol{\ell}^{(k)}=L(\tilde{\mathbf z}^{(k)}),\mathbf h^{(k)}=\tilde{\mathbf z}^{(k)}-\boldsymbol{\ell}^{(k)}$.

To provide a reference for alignment, we maintain momentum statistics computed only from pre-alignment low-frequency components. The momentum anchors are initialized with $\boldsymbol{\ell}^{(0)}$ and update after each turn according to the following rules:
\begin{equation}\mathbf m_{\mu}^{(k)}=\alpha_{\mu}\mathbf m_{\mu}^{(k-1)}+(1-\alpha_{\mu})\mu(\boldsymbol{\ell}^{(k)}),\quad \mathbf m_{\log\sigma}^{(k)}=\alpha_{\sigma}\mathbf m_{\log\sigma}^{(k-1)}+(1-\alpha_{\sigma})\log\sigma(\boldsymbol{\ell}^{(k)}),\end{equation}
where $\alpha_{\mu},\alpha_{\sigma}\in(0,1)$ are momentum coefficients. The momentum anchor stores exponential moving averages (EMA) of the mean and log-standard-deviation. Given these target statistics, VAE-LFA aligns the current low-frequency component channel-wise:
\vspace{-0.5mm}
\begin{equation}\boldsymbol{\ell}_{\mathrm{align},c}^{(k)}=
\mathbf m_{\mu,c}^{(k-1)}+\frac{\exp\Big(\mathbf m_{\log\sigma,c}^{(k-1)}\Big)}{\sigma_c\big(\boldsymbol{\ell}^{(k)}\big)+\epsilon}\left(\boldsymbol{\ell}_{c}^{(k)}-\mu_c\big(\boldsymbol{\ell}^{(k)}\big)\right),\quad c=1,\ldots,C.\end{equation}
The final latent is obtained by combining the aligned low-frequency component with the original high-frequency residual: $\hat{\mathbf z}^{(k)}=\boldsymbol{\ell}_{\mathrm{align}}^{(k)}+\mathbf h^{(k)}$.

\vspace{-2mm}
\subsection{The Implementation of VAE-LFA}
\vspace{-2mm}
\label{sec:implementation}

\paragraph{On white-box models.}
For open-source editors with full internal access, VAE-LFA is inserted \textbf{between consecutive DiT forward passes}, eliminating redundant VAE round trips. At turn $k$, the DiT output $\tilde{\mathbf{z}}^{(k)}$ is aligned into $\hat{\mathbf{z}}^{(k)}$, which is directly fed to the next DiT turn. The VAE decoder is used only to obtain the visible image $\mathbf{x}^{(k)}=D(\hat{\mathbf{z}}^{(k)})$, while the encoder is bypassed between turns. This reduces latency and removes VAE round-trip noise from the drift accumulation chain; the only persistent item is the EMA statistics.
\vspace{-0.1cm}

\paragraph{On black-box models.}
For API-only editors, we access only image-level inputs and outputs, without internal features. We interleave an off-the-shelf VAE $E_{\text{ext}},D_{\text{ext}}$ (e.g., SD-VAE-ft-ema~\cite{rombach2022high}) between rounds:
$\mathbf{x}^{(k-1)} \xrightarrow{\text{API}} \tilde{\mathbf{x}}^{(k)} \xrightarrow{E_{\text{ext}}} \tilde{\mathbf{z}}^{(k)} \xrightarrow{\text{VAE-LFA}} \hat{\mathbf{z}}^{(k)} \xrightarrow{D_{\text{ext}}} \mathbf{x}^{(k)}$.
Although the external VAE may differ from the editor's native latent space, VAE-LFA remains effective under cross-VAE deployment (Sec.~\ref{sec:experiments}), likely because it acts on low-frequency statistics. Since low-frequency components are comparatively robust to domain and architectural variations~\cite{yang2026structuredspectralreasoningfrequencyadaptive, ma2024decompositionbasedunsuperviseddomainadaptation, liu2024architecture}, external-VAE alignment can still suppress semantic drift. Implementation parameters are in Appendix~\ref{appendix:parameters}.
\vspace{-0.1cm}

\begin{figure*}[t]
  \centering
  \includegraphics[width=\linewidth]{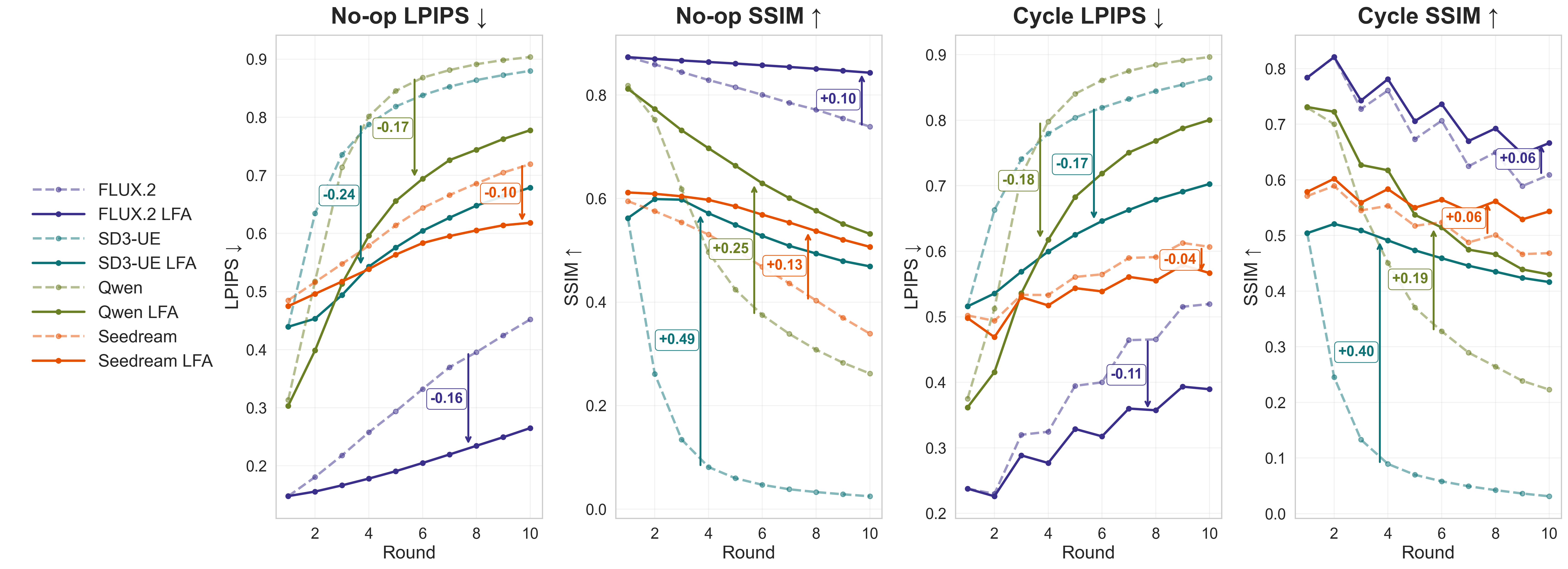}
  \vspace{-0.3cm}
  \caption{\textbf{Per-round metrics curve of no-op and cycle editing.} On average, VAE-LFA improves visual consistency both from a macro (LPIPS) and micro (SSIM) perspective.
  }
  \label{fig:rounds_curve}
  \vspace{-3mm}
\end{figure*}

\begin{table*}
  \caption{\textbf{No-op and cycle editing results for all models.} We abbreviate Qwen Image 2.0 as Qwen and Seedream 4.0 as Seedream. The better result between LFA/non-LFA variants is shown in \textbf{bold}.}
  \label{tab:noop_cycle}
  \centering
  \tiny
  \setlength{\tabcolsep}{1.3pt}
  \begin{tabular}{@{}p{2.05cm}*{18}{c}@{}}
    \toprule
    \multirow{3}{*}{Model} & \multicolumn{9}{c}{photograph} & \multicolumn{9}{c}{illustration} \\
    \cmidrule(lr){2-10} \cmidrule(lr){11-19}
    & \multicolumn{3}{c}{creature} & \multicolumn{3}{c}{architecture} & \multicolumn{3}{c}{scenery} & \multicolumn{3}{c}{creature} & \multicolumn{3}{c}{architecture} & \multicolumn{3}{c}{scenery} \\
    \cmidrule(lr){2-4} \cmidrule(lr){5-7} \cmidrule(lr){8-10} \cmidrule(lr){11-13} \cmidrule(lr){14-16} \cmidrule(lr){17-19}
    & LPIPS$\downarrow$ & L1$\downarrow$ & SSIM$\uparrow$ & LPIPS$\downarrow$ & L1$\downarrow$ & SSIM$\uparrow$ & LPIPS$\downarrow$ & L1$\downarrow$ & SSIM$\uparrow$ & LPIPS$\downarrow$ & L1$\downarrow$ & SSIM$\uparrow$ & LPIPS$\downarrow$ & L1$\downarrow$ & SSIM$\uparrow$ & LPIPS$\downarrow$ & L1$\downarrow$ & SSIM$\uparrow$ \\
    \midrule
    \multicolumn{19}{@{}l}{\textit{\textcolor{gray}{No-Op, Round 10}}} \\
    \multicolumn{19}{@{}l}{\textit{\textcolor{gray}{White-Box}}} \\
    FLUX.2 & 0.36 & 0.09 & 0.81 & 0.42 & 0.15 & 0.70 & 0.43 & 0.13 & 0.77 & 0.46 & 0.15 & 0.70 & 0.50 & 0.16 & 0.72 & 0.56 & 0.15 & 0.71 \\
    \prow \textbf{FLUX.2+LFA} & \textbf{0.17} & \textbf{0.04} & \textbf{0.93} & \textbf{0.27} & \textbf{0.09} & \textbf{0.79} & \textbf{0.25} & \textbf{0.07} & \textbf{0.83} & \textbf{0.26} & \textbf{0.07} & \textbf{0.88} & \textbf{0.29} & \textbf{0.09} & \textbf{0.80} & \textbf{0.33} & \textbf{0.08} & \textbf{0.83} \\
    SD3-UE & 0.87 & 0.42 & 0.03 & 0.90 & 0.48 & 0.01 & 0.88 & 0.41 & 0.01 & 0.82 & 0.46 & 0.08 & 0.89 & 0.52 & 0.02 & 0.92 & 0.44 & 0.01 \\
    \prow \textbf{SD3-UE+LFA} & \textbf{0.66} & \textbf{0.26} & \textbf{0.50} & \textbf{0.68} & \textbf{0.32} & \textbf{0.42} & \textbf{0.69} & \textbf{0.30} & \textbf{0.47} & \textbf{0.61} & \textbf{0.27} & \textbf{0.55} & \textbf{0.69} & \textbf{0.33} & \textbf{0.45} & \textbf{0.74} & \textbf{0.32} & \textbf{0.43} \\
    \midrule
    \multicolumn{19}{@{}l}{\textit{\textcolor{gray}{Black-Box}}} \\
    Qwen & 0.82 & 0.22 & 0.44 & 0.77 & 0.22 & 0.43 & 0.78 & 0.22 & 0.46 & 0.72 & 0.21 & 0.52 & 0.69 & 0.22 & 0.44 & 0.80 & 0.21 & 0.50 \\
    \prow \textbf{Qwen+LFA} & \textbf{0.66} & \textbf{0.12} & \textbf{0.68} & \textbf{0.64} & \textbf{0.15} & \textbf{0.62} & \textbf{0.60} & \textbf{0.11} & \textbf{0.71} & \textbf{0.57} & \textbf{0.14} & \textbf{0.66} & \textbf{0.54} & \textbf{0.16} & \textbf{0.60} & \textbf{0.68} & \textbf{0.15} & \textbf{0.67} \\
    Seedream & 0.66 & 0.15 & 0.54 & \textbf{0.61} & 0.19 & 0.42 & 0.61 & 0.18 & 0.48 & 0.64 & 0.20 & 0.45 & 0.59 & 0.21 & \textbf{0.42} & \textbf{0.61} & 0.17 & 0.56 \\
    \prow \textbf{Seedream+LFA} & \textbf{0.65} & \textbf{0.13} & \textbf{0.57} & 0.62 & \textbf{0.17} & \textbf{0.45} & 0.61 & \textbf{0.16} & \textbf{0.51} & \textbf{0.62} & \textbf{0.19} & \textbf{0.50} & 0.59 & \textbf{0.20} & 0.41 & 0.62 & \textbf{0.16} & \textbf{0.60} \\
    \midrule
    \multicolumn{19}{@{}l}{\textit{\textcolor{gray}{Cycle, Round 10}}} \\
    \multicolumn{19}{@{}l}{\textit{\textcolor{gray}{White-Box}}} \\
    FLUX.2 & 0.42 & 0.12 & 0.67 & 0.48 & 0.17 & 0.62 & 0.52 & 0.17 & 0.64 & 0.51 & 0.17 & 0.60 & 0.53 & 0.17 & 0.59 & 0.66 & 0.20 & 0.53 \\
    \prow \textbf{FLUX.2+LFA} & \textbf{0.30} & \textbf{0.08} & \textbf{0.73} & \textbf{0.31} & \textbf{0.12} & \textbf{0.67} & \textbf{0.40} & \textbf{0.13} & \textbf{0.67} & \textbf{0.31} & \textbf{0.09} & \textbf{0.70} & \textbf{0.40} & \textbf{0.13} & \textbf{0.63} & \textbf{0.53} & \textbf{0.16} & \textbf{0.58} \\
    SD3-UE & 0.84 & 0.41 & 0.03 & 0.88 & 0.48 & 0.01 & 0.89 & 0.42 & 0.01 & 0.81 & 0.45 & 0.10 & 0.86 & 0.51 & 0.03 & 0.90 & 0.44 & 0.01 \\
    \prow \textbf{SD3-UE+LFA} & \textbf{0.68} & \textbf{0.28} & \textbf{0.44} & \textbf{0.71} & \textbf{0.33} & \textbf{0.37} & \textbf{0.71} & \textbf{0.31} & \textbf{0.43} & \textbf{0.65} & \textbf{0.30} & \textbf{0.49} & \textbf{0.71} & \textbf{0.35} & \textbf{0.38} & \textbf{0.75} & \textbf{0.34} & \textbf{0.38} \\
    \midrule
    \multicolumn{19}{@{}l}{\textit{\textcolor{gray}{Black-Box}}} \\
    Qwen & 0.82 & 0.23 & 0.39 & 0.79 & 0.22 & 0.42 & 0.80 & 0.24 & 0.40 & 0.70 & 0.22 & 0.48 & 0.67 & 0.22 & 0.40 & 0.83 & 0.24 & 0.39 \\
    \prow \textbf{Qwen+LFA} & \textbf{0.69} & \textbf{0.16} & \textbf{0.53} & \textbf{0.65} & \textbf{0.15} & \textbf{0.57} & \textbf{0.66} & \textbf{0.17} & \textbf{0.56} & \textbf{0.57} & \textbf{0.15} & \textbf{0.62} & \textbf{0.55} & \textbf{0.17} & \textbf{0.54} & \textbf{0.74} & \textbf{0.19} & \textbf{0.50} \\
    Seedream & \textbf{0.61} & 0.16 & 0.51 & 0.57 & 0.18 & 0.43 & \textbf{0.61} & 0.19 & 0.47 & 0.63 & 0.21 & 0.43 & 0.58 & 0.22 & 0.43 & 0.64 & 0.18 & 0.54 \\
    \prow \textbf{Seedream+LFA} & 0.68 & \textbf{0.15} & \textbf{0.55} & \textbf{0.54} & \textbf{0.14} & \textbf{0.53} & 0.64 & \textbf{0.17} & \textbf{0.55} & \textbf{0.57} & \textbf{0.16} & \textbf{0.49} & \textbf{0.46} & \textbf{0.17} & \textbf{0.56} & \textbf{0.51} & \textbf{0.16} & \textbf{0.58} \\
    \bottomrule
  \end{tabular}
  \vspace{-1mm}
\end{table*}

\begin{table*}[h]
  \vspace{-1mm}
  \caption{\textbf{Long-chain editing results for all models.}}
  \label{tab:longchain}
  \centering
  \tiny
  \setlength{\tabcolsep}{1.5pt}
  \begin{tabular}{@{}p{2.05cm}*{16}{c}@{}}
    \toprule
    \multirow{4}{*}{Model} & \multicolumn{8}{c}{Round 5} & \multicolumn{8}{c}{Round 10} \\
    \cmidrule(lr){2-9} \cmidrule(lr){10-17}
    & \multicolumn{4}{c}{photograph} & \multicolumn{4}{c}{illustration} & \multicolumn{4}{c}{photograph} & \multicolumn{4}{c}{illustration} \\
    \cmidrule(lr){2-5} \cmidrule(lr){6-9} \cmidrule(lr){10-13} \cmidrule(lr){14-17}
    & \multicolumn{2}{c}{clear object} & \multicolumn{2}{c}{salient object} & \multicolumn{2}{c}{clear object} & \multicolumn{2}{c}{salient object} & \multicolumn{2}{c}{clear object} & \multicolumn{2}{c}{salient object} & \multicolumn{2}{c}{clear object} & \multicolumn{2}{c}{salient object} \\
    \cmidrule(lr){2-3} \cmidrule(lr){4-5} \cmidrule(lr){6-7} \cmidrule(lr){8-9} \cmidrule(lr){10-11} \cmidrule(lr){12-13} \cmidrule(lr){14-15} \cmidrule(lr){16-17}
    & DINO$\uparrow$ & VLM$\uparrow$ & DINO$\uparrow$ & VLM$\uparrow$ & DINO$\uparrow$ & VLM$\uparrow$ & DINO$\uparrow$ & VLM$\uparrow$ & DINO$\uparrow$ & VLM$\uparrow$ & DINO$\uparrow$ & VLM$\uparrow$ & DINO$\uparrow$ & VLM$\uparrow$ & DINO$\uparrow$ & VLM$\uparrow$ \\
    \midrule
    \multicolumn{17}{@{}l}{\textit{\textcolor{gray}{White-Box}}} \\
    FLUX.2 & 0.46 & 78.1 & 0.44 & 74.2 & 0.60 & 75.1 & 0.47 & 72.5 & 0.38 & 72.0 & 0.31 & 70.8 & 0.35 & \textbf{68.1} & 0.35 & 68.3 \\
    \prow \textbf{FLUX.2+LFA} & 0.46 & \textbf{78.4} & \textbf{0.48} & \textbf{75.6} & \textbf{0.62} & \textbf{75.8} & \textbf{0.48} & \textbf{73.2} & \textbf{0.40} & \textbf{74.4} & \textbf{0.34} & \textbf{71.7} & \textbf{0.54} & 67.0 & \textbf{0.37} & \textbf{71.0} \\
    SD3-UE & 0.22 & 63.6 & 0.20 & 59.0 & 0.19 & 56.6 & 0.18 & 47.3 & 0.11 & 53.2 & 0.11 & 48.8 & 0.27 & \textbf{50.0} & 0.13 & 47.6 \\
    \prow \textbf{SD3-UE+LFA} & \textbf{0.44} & \textbf{64.0} & \textbf{0.36} & \textbf{60.8} & \textbf{0.34} & \textbf{58.1} & \textbf{0.29} & \textbf{56.8} & \textbf{0.20} & \textbf{56.6} & \textbf{0.19} & \textbf{51.3} & \textbf{0.31} & 48.2 & \textbf{0.28} & \textbf{49.7} \\
    \midrule
    \multicolumn{17}{@{}l}{\textit{\textcolor{gray}{Black-Box}}} \\
    Qwen & 0.43 & 63.2 & 0.51 & 61.1 & 0.46 & 64.9 & 0.52 & 62.7 & 0.29 & 55.1 & 0.33 & 48.1 & 0.31 & 54.6 & 0.32 & 50.1 \\
    \prow \textbf{Qwen+LFA} & \textbf{0.49} & \textbf{66.7} & \textbf{0.55} & \textbf{62.1} & \textbf{0.51} & \textbf{68.3} & \textbf{0.54} & \textbf{65.6} & \textbf{0.37} & \textbf{60.9} & \textbf{0.44} & \textbf{53.8} & \textbf{0.39} & \textbf{58.7} & \textbf{0.41} & \textbf{55.2} \\
    Seedream & 0.50 & 72.1 & 0.52 & \textbf{74.3} & 0.48 & \textbf{76.6} & 0.53 & 74.5 & 0.37 & 70.0 & 0.35 & 67.7 & 0.32 & 68.3 & 0.41 & 69.0\\
    \prow \textbf{Seedream+LFA} & \textbf{0.61} & \textbf{73.0} & \textbf{0.56} & 72.6 & \textbf{0.60} & 74.7 & \textbf{0.59} & \textbf{75.9} & \textbf{0.44} & \textbf{70.6} & \textbf{0.36} & \textbf{68.0} & \textbf{0.42} & \textbf{69.4} & \textbf{0.47} & \textbf{70.1} \\
    \bottomrule
  \end{tabular}
  \vspace{-2mm}
\end{table*}

\begin{table*}
  % \vspace{-2mm}
  \caption{\textbf{FLUX score breakdown on long-chain editing.} IP, SC, VQ, and CP denote instruction progress, state consistency, visual quality, and content preservation. Best paired results are in \textbf{bold}.}
  \label{tab:flux_breakdown}
  \centering
  \tiny
  \setlength{\tabcolsep}{1.8pt}
  \resizebox{\textwidth}{!}{%
  \begin{tabular}{@{}ll*{20}{c}@{}}
    \toprule
    \multirow{3}{*}{Round} & \multirow{3}{*}{Model}
    & \multicolumn{10}{c}{Illustration}
    & \multicolumn{10}{c}{Photograph} \\
    \cmidrule(lr){3-12} \cmidrule(lr){13-22}
    & & \multicolumn{5}{c}{Clear Object} & \multicolumn{5}{c}{Salient Object}
      & \multicolumn{5}{c}{Clear Object} & \multicolumn{5}{c}{Salient Object} \\
    \cmidrule(lr){3-7} \cmidrule(lr){8-12} \cmidrule(lr){13-17} \cmidrule(lr){18-22}
    & & IP$\uparrow$ & SC$\uparrow$ & VQ$\uparrow$ & CP$\uparrow$ & Tot$\uparrow$
      & IP$\uparrow$ & SC$\uparrow$ & VQ$\uparrow$ & CP$\uparrow$ & Tot$\uparrow$
      & IP$\uparrow$ & SC$\uparrow$ & VQ$\uparrow$ & CP$\uparrow$ & Tot$\uparrow$
      & IP$\uparrow$ & SC$\uparrow$ & VQ$\uparrow$ & CP$\uparrow$ & Tot$\uparrow$ \\
    \midrule

    \multirow{2}{*}{R5}
    & FLUX.2
    & 28.5 & 16.4 & 14.1 & 16.1 & 75.1
    & \textbf{28.0} & 15.7 & 14.2 & 14.6 & 72.5
    & 30.2 & \textbf{17.0} & 14.8 & \textbf{16.1} & 78.1
    & 29.1 & 16.3 & \textbf{14.3} & 14.6 & 74.2 \\
    & \textbf{+LFA}
    & \textbf{29.2} & 16.4 & \textbf{14.7} & \textbf{16.2} & \textbf{75.8}
    & 27.3 & \textbf{16.2} & \textbf{14.7} & \textbf{15.0} & \textbf{73.2}
    & \textbf{31.0} & 16.4 & \textbf{15.2} & 15.8 & \textbf{78.4}
    & \textbf{29.6} & \textbf{16.8} & 14.1 & \textbf{15.1} & \textbf{75.6} \\

    \midrule

    \multirow{2}{*}{R10}
    & FLUX.2
    & \textbf{23.7} & 15.7 & 14.2 & 14.5 & \textbf{68.1}
    & 26.4 & \textbf{16.0} & 13.5 & 12.4 & 68.3
    & 28.9 & 15.2 & \textbf{13.6} & \textbf{14.4} & 72.0
    & 29.7 & 15.5 & \textbf{13.2} & 12.4 & 70.8 \\
    & \textbf{+LFA}
    & 22.0 & \textbf{15.9} & \textbf{14.3} & \textbf{14.8} & 67.0
    & \textbf{29.2} & 15.6 & \textbf{13.8} & 12.4 & \textbf{71.0}
    & \textbf{31.1} & \textbf{15.9} & 13.5 & 13.9 & \textbf{74.4}
    & \textbf{30.5} & 15.5 & 12.8 & \textbf{12.9} & \textbf{71.7} \\

    \bottomrule
  \end{tabular}%
  }
  \vspace{-3mm}
\end{table*}

\vspace{-0.3cm}
\section{Experiments}
\vspace{-0.1cm}
\label{sec:experiments}

\vspace{-1.5mm}
\subsection{Experiment Setup}
\vspace{-1.5mm}

\paragraph{Base models and data.}
We evaluate VAE-LFA on four DiT-based editors: two white-box models, FLUX.2 Klein 9B~\cite{flux-2-2025} and SD3-UE~\cite{esser2024scaling,zhao2024ultraeditinstructionbasedfinegrainedimage}, and two black-box API models, Qwen Image 2.0~\cite{wu2025qwen} and Seedream 4.0~\cite{seedream2025seedream}. For black-box editors, VAE-LFA uses an off-the-shelf SD-VAE-ft-ema~\cite{rombach2022high} between API calls, as described in Sec.~\ref{sec:implementation}. We evaluate on 120 images spanning illustration/photograph domains and creature/architecture/scenery categories; details are in Appendix~\ref{appendix:exp_details}.

\begin{figure*}[h]
  % \vspace{-2mm}
  \centering
  \includegraphics[width=0.9\linewidth]{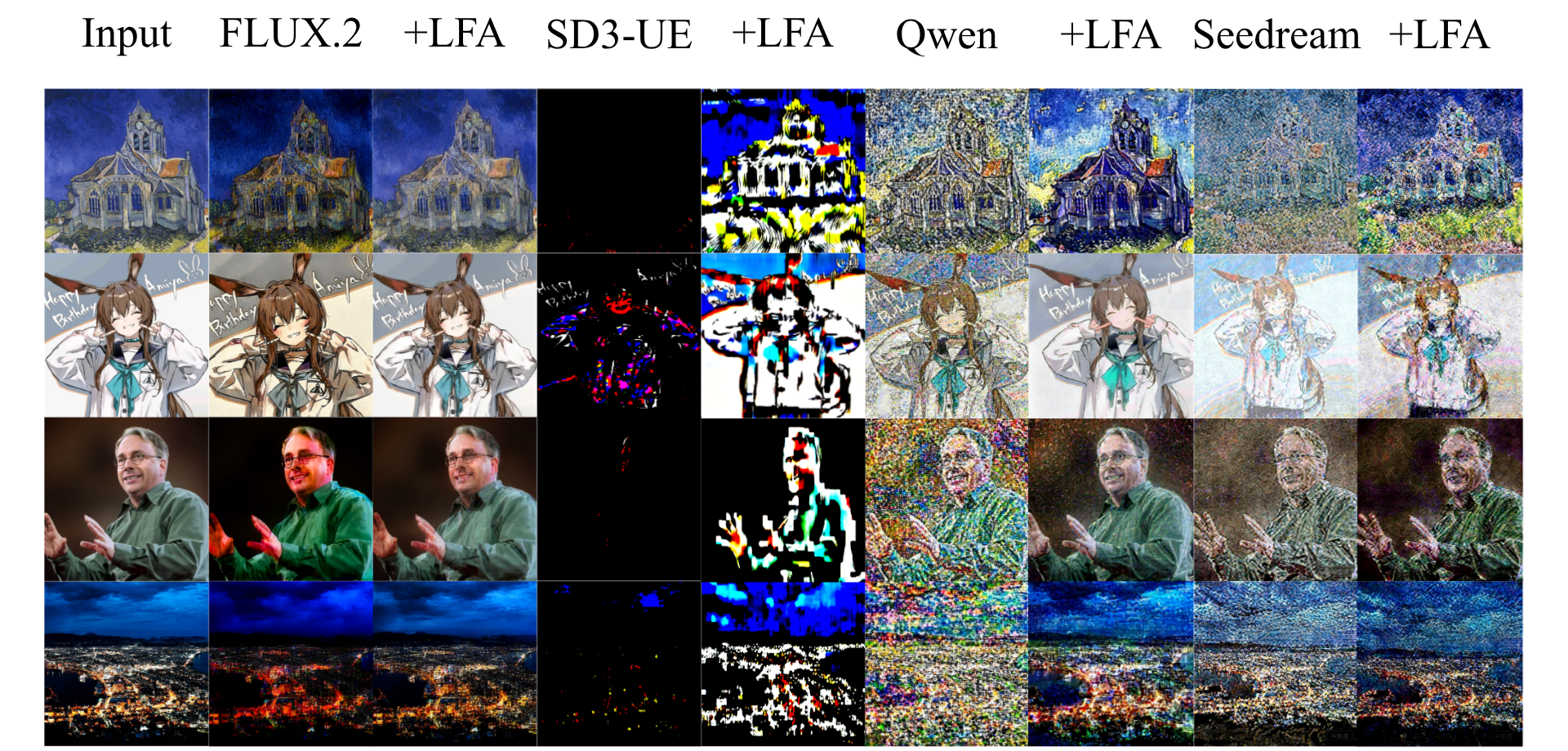}
  \vspace{-1mm}
  \caption{\textbf{Visualization of No-op editing trajectories at round 10.}
  }
  \label{fig:noop}
  \vspace{-3mm}
\end{figure*}

\begin{figure*}[h]
  \vspace{-2mm}
  \centering
  \includegraphics[width=0.9\linewidth]{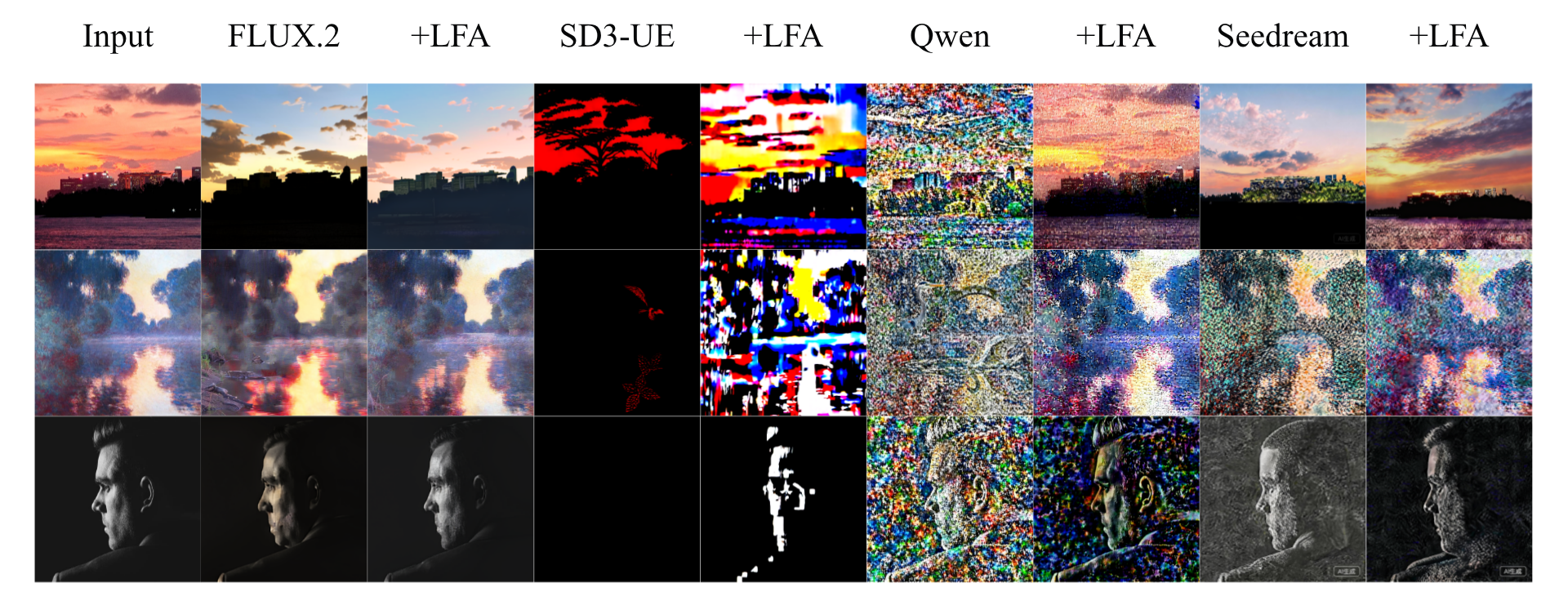}
  \vspace{-2mm}
  \caption{\textbf{Visualization of Cycle editing results at round 10.}
  }
  \label{fig:cycle}
  \vspace{-0.6cm}
\end{figure*}

\begin{figure*}
  \vspace{-3mm}
  \centering
  \includegraphics[width=0.9\linewidth]{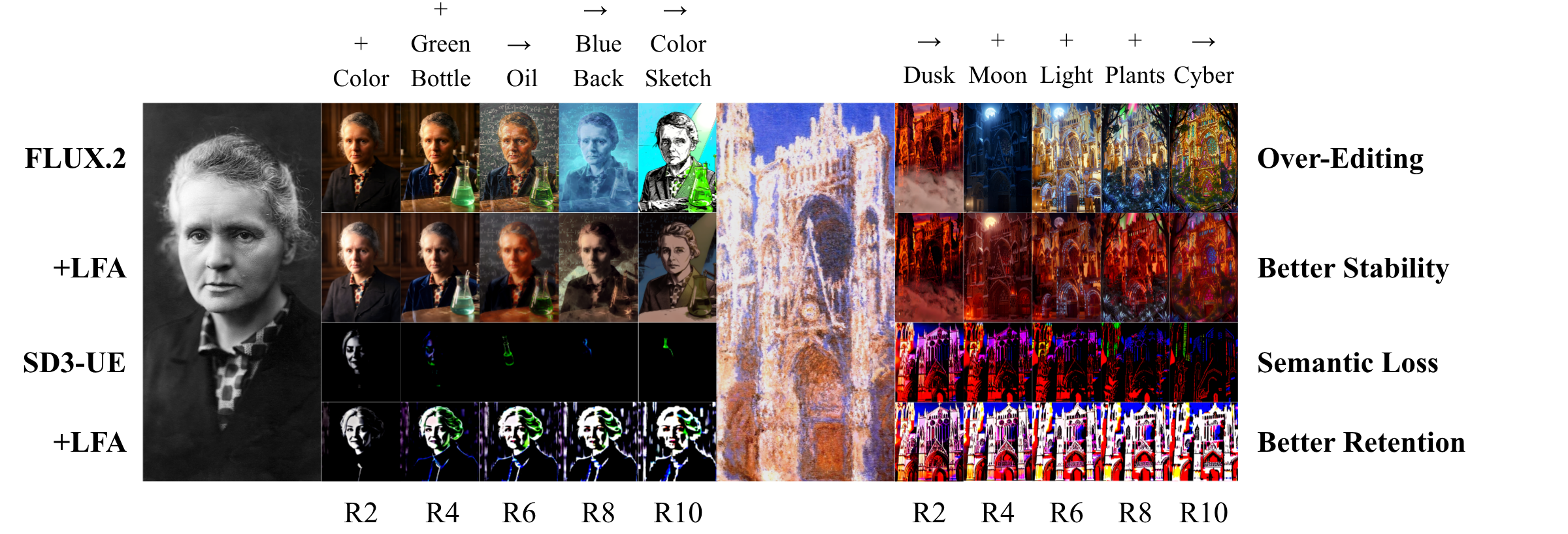}
  \caption{\textbf{Visualization of white-box long-chain edits at round 2, 4, 6, 8, 10.}
  }
  \label{fig:white_longchain}
  \vspace{-3mm}
\end{figure*}

\vspace{-2mm}
\paragraph{Evaluation protocol.}
Since VAE-LFA is a training-free post-hoc module, our primary comparison is paired: each editor is evaluated against its own +LFA variant under identical images, prompts, seeds, and model/API settings. This isolates the effect of inter-round low-frequency alignment without confounds from different training data, model capacity, or optimization recipes.

\vspace{-2mm}
\paragraph{Tasks and metrics.}
We use three 10-turn protocols. \textbf{No-op editing} applies identity-preserving prompts to isolate intrinsic drift; we report LPIPS~\cite{zhang2018unreasonableeffectivenessdeepfeatures}, SSIM~\cite{wang2004image}, and normalized L1 drift to the initial image. \textbf{Cycle editing} alternates inverse instructions and uses the same metrics. \textbf{Long-chain editing} applies cumulative edits and reports DINOv3~\cite{simeoni2025dinov3} subject consistency and Qwen-3.6-based~\cite{qwen36plus} VLM scores, following EdiVal-Agent~\cite{chen2025edival}. Implementation details are in Appendix~\ref{appendix:exp_details}--\ref{appendix:vlm_prompt}.

\begin{table*}[t]
\caption{\textbf{Ablation on frequency alignment scope for FLUX.2.} Best results are in \textbf{bold}. $^\ast$ uses the DiT-only setting; Low only is VAE-LFA. }
  \label{tab:ablation_freq}
  \centering
  \tiny
  \setlength{\tabcolsep}{1.5pt}
  \begin{tabular}{@{}l*{18}{c}@{}}
    \toprule
    \multirow{3}{*}{Variant} & \multicolumn{9}{c}{photograph} & \multicolumn{9}{c}{illustration} \\
    \cmidrule(lr){2-10} \cmidrule(lr){11-19}
    & \multicolumn{3}{c}{creature} & \multicolumn{3}{c}{architecture} & \multicolumn{3}{c}{scenery} & \multicolumn{3}{c}{creature} & \multicolumn{3}{c}{architecture} & \multicolumn{3}{c}{scenery} \\
    \cmidrule(lr){2-4} \cmidrule(lr){5-7} \cmidrule(lr){8-10} \cmidrule(lr){11-13} \cmidrule(lr){14-16} \cmidrule(lr){17-19}
    & LPIPS$\downarrow$ & L1$\downarrow$ & SSIM$\uparrow$ & LPIPS$\downarrow$ & L1$\downarrow$ & SSIM$\uparrow$ & LPIPS$\downarrow$ & L1$\downarrow$ & SSIM$\uparrow$ & LPIPS$\downarrow$ & L1$\downarrow$ & SSIM$\uparrow$ & LPIPS$\downarrow$ & L1$\downarrow$ & SSIM$\uparrow$ & LPIPS$\downarrow$ & L1$\downarrow$ & SSIM$\uparrow$ \\
    \midrule
    \multicolumn{19}{@{}l}{\textit{\textcolor{gray}{No-Op, Round 10}}} \\
    FLUX.2$^{*}$ (Base) & 0.26 & 0.07 & 0.90 & 0.36 & 0.13 & 0.78 & 0.33 & 0.11 & 0.81 & 0.28 & 0.09 & 0.84 & 0.33 & 0.11 & 0.78 & 0.38 & 0.12 & 0.77 \\
    + Both & \textbf{0.16} & 0.05 & \textbf{0.93} & \textbf{0.25} & \textbf{0.09} & \textbf{0.79} & \textbf{0.36} & \textbf{0.07} & \textbf{0.83} & \textbf{0.26} & \textbf{0.07} & \textbf{0.91} & \textbf{0.29} & \textbf{0.09} & 0.79 & \textbf{0.33} & \textbf{0.07} & \textbf{0.83} \\
    + High only & 0.22 & 0.07 & 0.90 & 0.32 & 0.12 & 0.78 & 0.31 & 0.11 & 0.82 & 0.28 & 0.08 & 0.87 & 0.31 & 0.11 & 0.77 & 0.24 & 0.11 & 0.79 \\
    \textbf{+ Low only (Ours)} & 0.17 & \textbf{0.04} & \textbf{0.93} & 0.27 & \textbf{0.09} & \textbf{0.79} & 0.25 & \textbf{0.07} & \textbf{0.83} & \textbf{0.26} & \textbf{0.07} & 0.88 & \textbf{0.29} & \textbf{0.09} & \textbf{0.80} & \textbf{0.33} & 0.08 & \textbf{0.83} \\
    \midrule
    \multicolumn{19}{@{}l}{\textit{\textcolor{gray}{Cycle, Round 10}}} \\
    FLUX.2$^{*}$ (Base) & 0.33 & 0.10 & 0.70 & 0.43 & 0.16 & 0.65 & 0.44 & 0.15 & 0.63 & \textbf{0.31} & 0.11 & 0.69 & 0.46 & 0.15 & 0.59 & 0.67 & 0.20 & 0.50 \\
    + Both & \textbf{0.30} & 0.09 & 0.71 & 0.33 & \textbf{0.12} & 0.65 & 0.41 & \textbf{0.13} & 0.65 & \textbf{0.31} & 0.10 & 0.69 & 0.42 & \textbf{0.13} & 0.61 & 0.55 & 0.18 & 0.56 \\
    + High only & 0.32 & 0.11 & 0.71 & 0.40 & 0.15 & 0.66 & 0.45 & 0.16 & 0.64 & 0.33 & 0.12 & 0.68 & 0.45 & 0.14 & 0.60 & 0.65 & 0.19 & 0.52 \\
    \textbf{+ Low only (Ours)} & \textbf{0.30} & \textbf{0.08} & \textbf{0.73} & \textbf{0.31} & \textbf{0.12} & \textbf{0.67} & \textbf{0.40} & \textbf{0.13} & \textbf{0.67} & \textbf{0.31} & \textbf{0.09} & \textbf{0.70} & \textbf{0.40} & \textbf{0.13} & \textbf{0.63} & \textbf{0.53} & \textbf{0.16} & \textbf{0.58} \\
    \bottomrule
  \end{tabular}
\end{table*}

\begin{table*}
\vspace{-1mm}
  \caption{\textbf{Ablation on anchor update strategy for FLUX.2.} EMA is the default VAE-LFA setting.}
  \label{tab:ablation_ema}
  \centering
  \tiny
  \setlength{\tabcolsep}{1.5pt}
  \begin{tabular}{@{}l*{18}{c}@{}}
    \toprule
    \multirow{3}{*}{Variant} & \multicolumn{9}{c}{photograph} & \multicolumn{9}{c}{illustration} \\
    \cmidrule(lr){2-10} \cmidrule(lr){11-19}
    & \multicolumn{3}{c}{creature} & \multicolumn{3}{c}{architecture} & \multicolumn{3}{c}{scenery} & \multicolumn{3}{c}{creature} & \multicolumn{3}{c}{architecture} & \multicolumn{3}{c}{scenery} \\
    \cmidrule(lr){2-4} \cmidrule(lr){5-7} \cmidrule(lr){8-10} \cmidrule(lr){11-13} \cmidrule(lr){14-16} \cmidrule(lr){17-19}
    & LPIPS$\downarrow$ & L1$\downarrow$ & SSIM$\uparrow$ & LPIPS$\downarrow$ & L1$\downarrow$ & SSIM$\uparrow$ & LPIPS$\downarrow$ & L1$\downarrow$ & SSIM$\uparrow$ & LPIPS$\downarrow$ & L1$\downarrow$ & SSIM$\uparrow$ & LPIPS$\downarrow$ & L1$\downarrow$ & SSIM$\uparrow$ & LPIPS$\downarrow$ & L1$\downarrow$ & SSIM$\uparrow$ \\
    \midrule
    \multicolumn{19}{@{}l}{\textit{\textcolor{gray}{No-Op, Round 10}}} \\
    FLUX.2$^{*}$ (Base) & 0.26 & 0.07 & 0.90 & 0.36 & 0.13 & 0.78 & 0.33 & 0.11 & 0.81 & 0.28 & 0.09 & 0.84 & 0.33 & 0.11 & 0.78 & 0.38 & 0.12 & 0.77 \\
    + Fixed & 0.18 & 0.05 & \textbf{0.93} & 0.29 & \textbf{0.09} & \textbf{0.79} & 0.26 & 0.08 & 0.82 & \textbf{0.26} & \textbf{0.07} & 0.86 & 0.30 & \textbf{0.09} & 0.79 & \textbf{0.33} & 0.09 & \textbf{0.83} \\
    + Prev & \textbf{0.17} & \textbf{0.04} & 0.92 & 0.28 & \textbf{0.09} & \textbf{0.79} & \textbf{0.25} & \textbf{0.07} & 0.84 & 0.27 & 0.08 & 0.87 & \textbf{0.29} & 0.10 & 0.79 & 0.35 & 0.09 & 0.81 \\
    \textbf{+ EMA (Ours)} & \textbf{0.17} & \textbf{0.04} & \textbf{0.93} & \textbf{0.27} & \textbf{0.09} & \textbf{0.79} & \textbf{0.25} & \textbf{0.07} & \textbf{0.83} & \textbf{0.26} & \textbf{0.07} & \textbf{0.88} & \textbf{0.29} & \textbf{0.09} & \textbf{0.80} & \textbf{0.33} & \textbf{0.08} & \textbf{0.83} \\
    \midrule
    \multicolumn{19}{@{}l}{\textit{\textcolor{gray}{Cycle, Round 10}}} \\
    FLUX.2$^{*}$ (Base) & 0.33 & 0.10 & 0.70 & 0.43 & 0.16 & 0.65 & 0.44 & 0.15 & 0.63 & \textbf{0.31} & 0.11 & 0.69 & 0.46 & 0.15 & 0.59 & 0.67 & 0.20 & 0.50 \\
    + Fixed & 0.32 & 0.09 & 0.72 & 0.32 & 0.13 & 0.66 & 0.41 & 0.13 & 0.65 & \textbf{0.31} & 0.10 & \textbf{0.70} & 0.42 & \textbf{0.13} & \textbf{0.62} & 0.56 & 0.17 & 0.54 \\
    + Prev & 0.31 & 0.09 & 0.71 & 0.34 & 0.14 & \textbf{0.67} & 0.42 & 0.14 & 0.65 & \textbf{0.31} & 0.11 & 0.69 & 0.43 & 0.14 & \textbf{0.62} & 0.59 & 0.18 & 0.53 \\
    \textbf{+ EMA (Ours)} & \textbf{0.30} & \textbf{0.08} & \textbf{0.73} & \textbf{0.31} & \textbf{0.12} & \textbf{0.67} & \textbf{0.40} & \textbf{0.13} & \textbf{0.67} & \textbf{0.31} & \textbf{0.09} & \textbf{0.70} & \textbf{0.40} & \textbf{0.13} & \textbf{0.63} & \textbf{0.53} & \textbf{0.16} & \textbf{0.58} \\
    \bottomrule
  \end{tabular}
  \vspace{-3mm}
\end{table*}

\vspace{-2mm}
\subsection{Quantitative Analysis}
\vspace{-2mm}

Fig.~\ref{fig:rounds_curve} and Tab.~\ref{tab:noop_cycle} report 10-round no-op and cycle editing results. VAE-LFA improves consistency metrics across both white-box and black-box editors. Tab.~\ref{tab:longchain} reports long-chain results, and Tab.~\ref{tab:flux_breakdown} provides the FLUX.2 VLM score breakdown. For long-chain edits, we group samples by clear-object and salient-object images (Appendix~\ref{appendix:salient_clear}), since object presence is more relevant than the original category. VAE-LFA improves consistency, instruction following, and overall visual quality on average. Bootstrap confidence intervals for no-op and cycle improvements are reported in Appendix~\ref{appendix:stats}.

\vspace{-2mm}
\subsection{Qualitative Analysis}
\vspace{-0.1cm}
Fig.~\ref{fig:noop}, Fig.~\ref{fig:cycle}, and Fig.~\ref{fig:white_longchain} show qualitative results for the three settings. Baseline editors accumulate color shifts, artifacts, and semantic leakage over turns. VAE-LFA better preserves appearance and structure without visibly degrading instruction following. More examples are provided in Appendix~\ref{appendix:qual}.

\vspace{-3mm}
\subsection{Ablation Studies}
\vspace{-2mm}
\paragraph{Frequency scope.}
Tab.~\ref{tab:ablation_freq} compares aligning low-frequency, high-frequency, and both components on FLUX.2. Low-frequency alignment achieves nearly the same no-op gains as full alignment, supporting our observation that drift is mainly low-frequency. High-frequency alignment helps little, while full alignment can over-constrain valid edits in the cycle setting. We therefore use low-frequency-only alignment by default.
\vspace{-2mm}
\paragraph{Anchor update.}
Tab.~\ref{tab:ablation_ema} compares fixed, previous-round, and EMA anchors for low-frequency alignment. The fixed anchor preserves the initial appearance but is rigid for meaningful edits, while the previous-round anchor can drift with the trajectory. EMA offers a smoother historical reference that balances stability and edit flexibility, and is used by default.
\vspace{-2mm}
\paragraph{External VAE.}
For black-box APIs, VAE-LFA introduces an external VAE encode--decode round trip, a potential confounder. We therefore include a +VAE baseline in Appendix~\ref{appendix:black_box_ablation}, using the same VAE round trip without low-frequency alignment. Its limited and inconsistent gains suggest that black-box improvements mainly come from VAE-LFA rather than VAE re-encoding alone.

% \vspace{-0.1cm}

\vspace{-3mm}
\section{Conclusion}
\label{sec:conclusion}
\vspace{-3mm}
We analyze multi-turn degradation in DiT-based image editors from a VAE-latent frequency perspective, showing that semantic drift is mainly driven by DiT-induced low-frequency deviations. Motivated by this, we propose VAE-LFA, a training-free low-frequency alignment method for both white-box and black-box editors. Experiments across no-op, cycle, and long-chain editing show improved consistency and fidelity.

\vspace{-2mm}
\paragraph{Discussion and limitations.}
% \label{sec:discussion}
% \vspace{-2mm}
VAE-LFA is effective in most editing scenarios, but its low-frequency alignment can mildly suppress intended large global changes under strong prompts like \texttt{``Change the black image into a fully white image.''} A possible remedy is to adapt the alignment strength according to prompt intensity. We also observe faster degradation under no-op editing, suggesting that multi-turn identity-preserving prompts could be useful in post-training curricula.
\begin{ack}
We thank our colleagues and collaborators for insightful discussions, careful feedback, and support throughout this project. We are grateful to the developers and maintainers of the open-source models, libraries, and evaluation tools that made our experiments possible. We also thank the broader generative modeling and image editing communities for building the foundations on which this work is based. We hope that we have provided meaningful insight into the future development of image generation and editing models.

\paragraph{Funding disclosure.}
The authors received no specific funding for this work.

\end{ack}
\clearpage
\bibliographystyle{plain}
\bibliography{bibliography}

%%%%%%%%%%%%%%%%%%%%%%%%%%%%%%%%%%%%%%%%%%%%%%%%%%%%%%%%%%%%

\appendix

% \section{Additional Formulation}
\newpage
\section{Impact Statement}
\label{appendix:impact}

VAE-LFA aims to improve the stability and consistency of multi-turn image editing, which can benefit creative workflows, visual design, content production, and iterative human-AI collaboration. By reducing unintended semantic drift and quality degradation, the method may make image editing systems more controllable and predictable for users.

At the same time, improved multi-turn editing consistency may also increase the realism and persistence of manipulated visual content, including misleading edits, impersonation, or other forms of synthetic media misuse. VAE-LFA does not introduce a new image generation model or bypass the safety mechanisms of existing editors, but it can be composed with such systems. Therefore, deployment should rely on the safeguards of the underlying image editing platforms, such as content provenance, watermarking, misuse detection, safety filtering, and policy-based restrictions for harmful editing requests.

\section{Detailed Related Works}
\label{appendix:related_works}

\paragraph{Instruction-based and multi-turn image editing.}
Instruction-based image editing aims to modify an input image according to natural-language instructions while preserving irrelevant content. Early and recent systems improve single-turn editing through paired edit data, manually annotated instruction-edit datasets, multimodal instruction enhancement, and stronger generative backbones~\cite{brooks2023instructpix2pix, zhang2023magicbrush, fu2023guiding, geng2024instructdiffusion, huang2024smartedit}. More recently, diffusion-transformer and flow-matching based editors have further advanced prompt following, realism, and in-context editing capabilities~\cite{labs2025flux, wu2025qwen, seedream2025seedream, glmimage}. However, these systems are still commonly optimized and evaluated in one-shot settings. When applied iteratively, small deviations introduced at each turn can accumulate, resulting in identity drift, color shift, oversmoothing, artifact amplification, or unintended semantic changes. Recent works have begun to study this multi-turn setting explicitly~\cite{zhou2025multi, liao2025freqedit, kim2025improving}, but robust iterative editing remains challenging, especially when the underlying editor is accessible only through an API.

\paragraph{Remedies for iterative degradation.}
Existing remedies for multi-turn degradation differ in the type of access and supervision they require. Training-based methods improve iterative or in-context editing by learning from richer supervision. For example, VINCIE learns in-context image editing from videos using interleaved multimodal sequences and a diffusion-transformer architecture~\cite{qu2025vincie}, while Emu Edit improves instruction-based editing through multi-task training over recognition and generation tasks~\cite{sheynin2024emu}. Other works target degradation mechanisms more directly. REED-VAE retrains the VAE to reduce artifacts and noise accumulated through repeated transitions between pixel and latent spaces~\cite{almog2025reed}. Layer-wise Memory stores internal latent representations and prompt embeddings from previous editing steps to maintain consistency during sequential modification~\cite{kim2025improving}. FreqEdit identifies progressive high-frequency information loss as a cause of multi-turn quality degradation and injects high-frequency features from reference velocity fields to stabilize editing~\cite{liao2025freqedit}. These approaches are complementary to ours, but they typically require additional training, access to internal representations or velocity fields, or assumptions about the editor's inference process. In contrast, our goal is to design a minimal post-hoc intervention that can operate at the VAE latent interface and remain applicable to both white-box and black-box editors.

\paragraph{VAE bottlenecks and latent-space analysis.}
Most modern latent diffusion and DiT-based image generators employ an autoencoder bottleneck that maps images into a lower-dimensional latent space before generative modeling~\cite{rombach2022high, kingma2013auto}. This design substantially reduces the computational cost of diffusion modeling compared with operating directly in pixel space~\cite{rombach2022high}. At the same time, the autoencoder is not an identity map: its encode--decode process can introduce reconstruction bias, and repeated transitions between pixel and latent spaces have been observed to accumulate artifacts and noise in iterative editing~\cite{almog2025reed}. Recent studies further show that the structure of the latent space affects generative modeling, for example through equivariance properties of autoencoder latents and through latent diffusability under diffusion training~\cite{kouzelis2025eqvaeequivarianceregularizedlatent, gu2026understandinglatentdiffusabilityfisher}. These works motivate analyzing the VAE latent space, but they do not directly attribute multi-turn editing drift to different stages of a VAE-DiT editing pipeline. In contrast, we use the VAE latent space as a shared interface to separate the DiT transition from the VAE round trip. Our controlled no-operation analysis shows that the VAE mainly contributes a comparatively stable reconstruction bias, whereas progressive semantic drift is dominated by low-frequency deviations introduced by the DiT.

\paragraph{Frequency perspectives in generative models.}
Frequency-domain analysis provides a useful way to inspect how generative models distribute information across spatial scales. Spectrum Matching studies VAE latent diffusability from a spectral perspective, arguing that latent power spectra and frequency-to-frequency decoder correspondence are important for latent diffusion quality~\cite{ning2026spectrummatchingunifiedperspective}. Other work on latent tokenizers shows that fine visual details are strongly tied to high-frequency fidelity, and that common reconstruction objectives may prioritize low-frequency reconstruction at the expense of high-frequency details~\cite{medi2025missingfinedetailsimages}. Frequency separation has also been used in generation architectures, where low-frequency components are associated with global semantics and high-frequency components with local detail synthesis~\cite{ma2025deco, zhao2026luve}.

Frequency-aware editing methods further show that manipulating different spectral bands can improve edit controllability. FreeDiff truncates excessive low-frequency guidance during diffusion sampling to reduce editing spillover~\cite{wu2024freediffprogressivefrequencytruncation}, while FlexiEdit refines high-frequency DDIM latent components to better support non-rigid edits~\cite{koo2024flexieditfrequencyawarelatentrefinement}. Beyond editing, DERO uses frequency-domain analysis of diffusion denoising to improve watermark robustness against diffusion-model erasure~\cite{fang2024dero}. These studies support the utility of frequency analysis, but they do not attribute multi-turn drift to VAE versus DiT components or design a black-box-compatible stabilization module.

Works on multimodal LLMs~\cite{tu2026smspplugandplaystrategymultiscale} and DiT latent semantic editing~\cite{shuai2024latent} also indicate that latent representations can expose controllable semantic structure. In contrast, we analyze editing bias directly in the VAE latent space by decomposing each latent channel into low- and high-frequency components. Our experiments show that the DiT-induced drift is concentrated in low-frequency latent components, while high-frequency residuals are better left unconstrained to preserve local details and editing flexibility. This observation motivates VAE-LFA: after each DiT transition, we align only the channel-wise mean and standard deviation of the low-frequency latent component to an EMA anchor, and then recombine it with the original high-frequency residual.

\section{Detailed Formulation of Multi-turn Editing Bias}
\label{appendix:formulation}
\vspace{-0.1cm}
Here we formulate semantic drift in VAE-DiT models during multi-turn editing. For simplicity, we ignore the stochasticity of the generative process and present a deterministic formulation.
\vspace{-0.1cm}

\paragraph{Image-space multi-turn drift.}
Let $\mathcal X$ denote the image space and let $\mathbf{x}^{(k)}\in\mathcal X$ be the image after the $k$-th editing turn. A multi-turn editor induces a sequence
$\mathbf{x}^{(k)}=\mathcal T_{p^{(k)}}(\mathbf{x}^{(k-1)})$, where $p^{(k)}$ is the instruction at turn $k$ and $\mathcal T_{p^{(k)}}:\mathcal X\rightarrow\mathcal X$ is the image-space editing transition. Let $\mathcal P_{\mathrm{no\text{-}op}}$ denote a set of no-operation prompts, such as ``keep the image unchanged'', and let $p_{\mathrm{no\text{-}op}}\in\mathcal P_{\mathrm{no\text{-}op}}$ be a representative prompt. Ideally, no-operation editing should be an identity map, i.e., $\mathcal T_{p_{\mathrm{no\text{-}op}}}(\mathbf{x})=\mathbf{x}$ for any $\mathbf{x}\in\mathcal X$. In practice, however, repeated no-operation editing violates this fixed point and produces visible degradation, including color noise, whitening, detail loss, and semantic ambiguity. Such degradation becomes increasingly pronounced as the number of editing turns increases. We refer to this no-operation fixed-point violation as \emph{multi-turn editing bias}. This definition distinguishes degradation from valid edits, since meaningful editing is expected to change the image.
\vspace{-0.1cm}
%\paragraph{Image space multi-turn drift.}Let $\mathcal X$ denote the image space and let $\mathbf{x}^{(k)}\in\mathcal X$ be the image after the $k$-th editing turn, with $\mathbf{x}^{(0)}$ denoting the initial input. Let $p^{(k)}$ be the prompt at the $k$-th editing turn and let $\mathcal T_{p^{(k)}}:\mathcal X\rightarrow\mathcal X$ denote the image space transition induced by the editor. A $K$-turn editing sequence can be formulated by$\mathbf{x}^{(k)} = \mathcal T_{p^{(k)}}\!\big(\mathbf{x}^{(k-1)}\big)$, $k=1,\ldots,K.$ Let $\mathcal P_{\mathrm{no\text{-}op}}$ denote a set of no-operation prompts, e.g., ``keep the image unchanged''. Ideally, an editor preserves the input image under no-operation prompts, which means for all $p_{\mathrm{no\text{-}op}}\in \mathcal P_{\mathrm{no\text{-}op}}$, $\mathcal T_{p_{\mathrm{no\text{-}op}}}(\mathbf{x}) = \mathbf{x}$ holds for all $x\in \mathcal{X}$. %Therefore, under repeated no-operation editing, an ideal trajectory should satisfy $\mathbf{x}^{(k)} = \mathbf{x}^{(0)}$ for all $k=1,\ldots,K$.

%In practice, however, repeated no-operation editing produces observable degradation, such as color noise, whitening, detail loss, and semantic ambiguity. We refer to this violation of the identity fixed point as \emph{multi-turn editing bias}. This is not to be confused with normal image edits, since meaningful editing is expected to modify the image.
\paragraph{Latent space transition.}
We analyze this bias in the VAE latent space. Let $E:\mathcal X\rightarrow\mathcal Z$ and $D:\mathcal Z\rightarrow\mathcal X$ denote the VAE encoder and decoder. Given an input latent $\mathbf z$, the prompt-conditioned DiT editor produces a raw latent $G_p(\mathbf z)$, and the standard VAE-DiT pipeline maps it to the next latent by $\Phi_p(\mathbf z):=E\!\left(D\!\left(G_p(\mathbf z)\right)\right)$. Analogous to the pixel space transition, for any no-operation prompt $p_{\mathrm{no\text{-}op}}$, $\Phi_{p_{\mathrm{no\text{-}op}}}$ should be an identity map ideally, such that $\Phi_{p_{\mathrm{no\text{-}op}}}(\mathbf{z})=\mathbf{z}$. Its violation defines the latent no-operation bias, which can be decomposed into two terms as:
\vspace{-0.1cm}
\begin{equation}
\begin{aligned}
\Phi_{p_{\mathrm{no\text{-}op}}}(\mathbf{z})-\mathbf{z} =\underbrace{G_{p_{\mathrm{no\text{-}op}}}(\mathbf{z})-\mathbf{z}}_{\text{DiT no-operation bias}} + \underbrace{\Phi_{p_{\mathrm{no\text{-}op}}}(\mathbf{z})-G_{p_{\mathrm{no\text{-}op}}}(\mathbf{z})}_{\text{VAE round-trip bias}}
\end{aligned}
\end{equation}

\vspace{-0.1cm}
\paragraph{Prompted editing as bias-contaminated displacement.}
For a meaningful prompt $p$, the one-step latent displacement contains both prompt-induced change and the no-operation bias:
\begin{equation}\Phi_p(\mathbf{z})-\mathbf{z}=\underbrace{\Phi_p(\mathbf{z})-\Phi_{p_{\mathrm{no\text{-}op}}}(\mathbf{z})}_{\text{prompt-induced displacement}}+\underbrace{\Phi_{p_{\mathrm{no\text{-}op}}}(\mathbf{z})-\mathbf{z}}_{\text{no-operation bias}} \label{eq:prompt_bias}\end{equation}
In prompted editing, the instruction-induced change and the model's intrinsic distribution drift are coupled through the same latent state and editing dynamics. The bracketed term is not guaranteed to be a clean semantic edit, and Eq.~\eqref{eq:prompt_bias} should not be interpreted as an independent additive decomposition. We use the no-operation transition only as a reference to expose the drift that exists even without a semantic edit. Although two terms are not independent, later experiments empirically verify that reducing this no-operation drift also mitigates degradation under meaningful prompts.
\vspace{-0.1cm}
\paragraph{Multi-turn accumulation.}
Total latent displacement over $K$ turns is an accumulation of one-step residuals. Each residual is evaluated at the current latent state, which already contains the effects of previous rounds. Thus, even small per-turn no-operation biases can accumulate over long trajectories.
\vspace{-0.1cm}
\vspace{-0.3cm}

\section{Additional Quantitative Results}

\subsection{Experiment Statistical Significance}
\label{appendix:stats}

All no-op and cycle comparisons are paired over identical images and instruction sequences, so each difference is computed between a base editor and its corresponding +LFA variant on the same sample. To quantify uncertainty, we report non-parametric percentile bootstrap 95\% confidence intervals (CIs) for the aggregated round-10 improvements in Tab.~\ref{tab:bootstrap_ci}. For LPIPS and L1, where lower is better, the paired improvement is defined as $\Delta_i = s_i^{\mathrm{base}} - s_i^{\mathrm{+LFA}}$; for SSIM, where higher is better, it is defined as $\Delta_i = s_i^{\mathrm{+LFA}} - s_i^{\mathrm{base}}$. Positive values therefore indicate improvement.

For each model and protocol, we first average the round-10 metric across two repeated runs for every sample, and then compute paired differences. We pool paired differences from FLUX.2 ($n=120$), SD3-UE ($n=120$), Qwen Image 2.0 ($n=120$), and Seedream 4.0 ($n=120$), yielding $N=480$ paired observations per protocol. We draw $B=10{,}000$ bootstrap resamples of size $N$ with replacement, compute the mean improvement for each resample, and use the 2.5-th and 97.5-th percentiles as the CI bounds. This percentile bootstrap procedure is non-parametric and does not assume normality of the paired differences.

\begin{table}[t]
\centering
\caption{Bootstrap 95\% confidence intervals of paired improvements of +LFA over base editors at round 10. Positive values indicate improvement.}
\vspace{5mm}
\label{tab:bootstrap_ci}
\begin{tabular}{lccc}
\toprule
Protocol & LPIPS $\uparrow$ & L1 $\uparrow$ & SSIM $\uparrow$ \\
\midrule
No-op & $0.1363$ [$0.1139$, $0.1487$] & $0.0783$ [$0.0458$, $0.0909$] & $0.1929$ [$0.1486$, $0.2272$] \\
Cycle & $0.1179$ [$0.1036$, $0.1292$] & $0.0692$ [$0.0563$, $0.0817$] & $0.1633$ [$0.1302$, $0.1969$] \\
\bottomrule
\end{tabular}
\end{table}

All reported confidence intervals in Tab.~\ref{tab:bootstrap_ci} are strictly positive, indicating that +LFA yields statistically significant paired improvements over the corresponding base editors for both no-op and cycle protocols at the 95\% confidence level. These results support the reliability of the consistency gains reported in the main tables.

\paragraph{Data preparation.} For each model–protocol combination, per-sample metrics (LPIPS, L1, SSIM) are extracted from the round-10 records of all available samples. Each condition (model, protocol, VAE status) was run twice; when a sample appears in multiple runs, its round-10 values are averaged to obtain a single observation per sample.

\subsection{Additional Ablation on Black-Box Models}
\label{appendix:black_box_ablation}

We provide an additional ablation experiment on black-box models, where we compare VAE-LFA with an external-VAE-only baseline. As shown in Tab.~\ref{tab:vae_ablation}, editing consistency is not consistently improved when applying an external VAE only, illustrating the effectiveness of VAE-LFA.

\begin{table*}
  \caption{Ablation studies on black-box models. The best result among variants of each base editor is shown in \textbf{bold}.}
  \label{tab:vae_ablation}
  \centering
  \tiny
  \setlength{\tabcolsep}{1.5pt}
  \begin{tabular}{@{}l*{18}{c}@{}}
    \toprule
    \multirow{3}{*}{Model} & \multicolumn{9}{c}{photograph} & \multicolumn{9}{c}{illustration} \\
    \cmidrule(lr){2-10} \cmidrule(lr){11-19}
    & \multicolumn{3}{c}{creature} & \multicolumn{3}{c}{architecture} & \multicolumn{3}{c}{scenery} & \multicolumn{3}{c}{creature} & \multicolumn{3}{c}{architecture} & \multicolumn{3}{c}{scenery} \\
    \cmidrule(lr){2-4} \cmidrule(lr){5-7} \cmidrule(lr){8-10} \cmidrule(lr){11-13} \cmidrule(lr){14-16} \cmidrule(lr){17-19}
    & LPIPS$\downarrow$ & L1$\downarrow$ & SSIM$\uparrow$ & LPIPS$\downarrow$ & L1$\downarrow$ & SSIM$\uparrow$ & LPIPS$\downarrow$ & L1$\downarrow$ & SSIM$\uparrow$ & LPIPS$\downarrow$ & L1$\downarrow$ & SSIM$\uparrow$ & LPIPS$\downarrow$ & L1$\downarrow$ & SSIM$\uparrow$ & LPIPS$\downarrow$ & L1$\downarrow$ & SSIM$\uparrow$ \\
    \midrule
    \multicolumn{19}{@{}l}{\textit{\textcolor{gray}{No-Op, Round 10}}} \\
    Qwen & 0.82 & 0.22 & 0.44 & 0.77 & 0.22 & 0.43 & 0.78 & 0.22 & 0.46 & 0.72 & 0.21 & 0.52 & 0.69 & 0.22 & 0.44 & 0.80 & 0.21 & 0.50 \\
    Qwen+VAE & 0.80 & 0.19 & 0.48 & 0.73 & 0.21 & 0.46 & 0.75 & 0.22 & 0.47 & 0.69 & 0.19 & 0.54 & 0.65 & 0.18 & 0.49 & 0.78 & 0.20 & 0.52 \\
    \textbf{Qwen+LFA} & \textbf{0.66} & \textbf{0.12} & \textbf{0.68} & \textbf{0.64} & \textbf{0.15} & \textbf{0.62} & \textbf{0.60} & \textbf{0.11} & \textbf{0.71} & \textbf{0.57} & \textbf{0.14} & \textbf{0.66} & \textbf{0.54} & \textbf{0.16} & \textbf{0.60} & \textbf{0.68} & \textbf{0.15} & \textbf{0.67} \\
    Seedream & 0.66 & 0.15 & 0.54 & \textbf{0.61} & 0.19 & 0.42 & \textbf{0.61} & 0.18 & 0.48 & 0.64 & 0.20 & 0.45 & \textbf{0.59} & 0.21 & \textbf{0.42} & \textbf{0.61} & 0.17 & 0.56 \\
    Seedream+VAE & 0.68 & 0.16 & 0.54 & 0.64 & 0.20 & 0.39 & 0.63 & 0.21 & 0.46 & 0.64 & 0.21 & 0.48 & 0.62 & 0.23 & 0.39 & 0.64 & 0.19 & 0.54 \\
    Seedream+LFA & \textbf{0.65} & \textbf{0.13} & \textbf{0.57}
    & 0.62 & \textbf{0.17} & \textbf{0.45}
    & \textbf{0.61} & \textbf{0.16} & \textbf{0.51}
    & \textbf{0.62} & \textbf{0.19} & \textbf{0.50}
    & \textbf{0.59} & \textbf{0.20} & 0.41
    & 0.62 & \textbf{0.16} & \textbf{0.60} \\
    \bottomrule
  \end{tabular}
\end{table*}

\section{Additional Qualitative Results}
\label{appendix:qual}
% We provide more qualitative examples here to better illustrate the visual effects of VAE-LFA.

% \subsection{No-Op Editing}

% \subsection{Cycle Editing}

% \subsection{Long-Chain Editing}

\begin{figure*}[t]
  \centering
  \includegraphics[width=\linewidth]{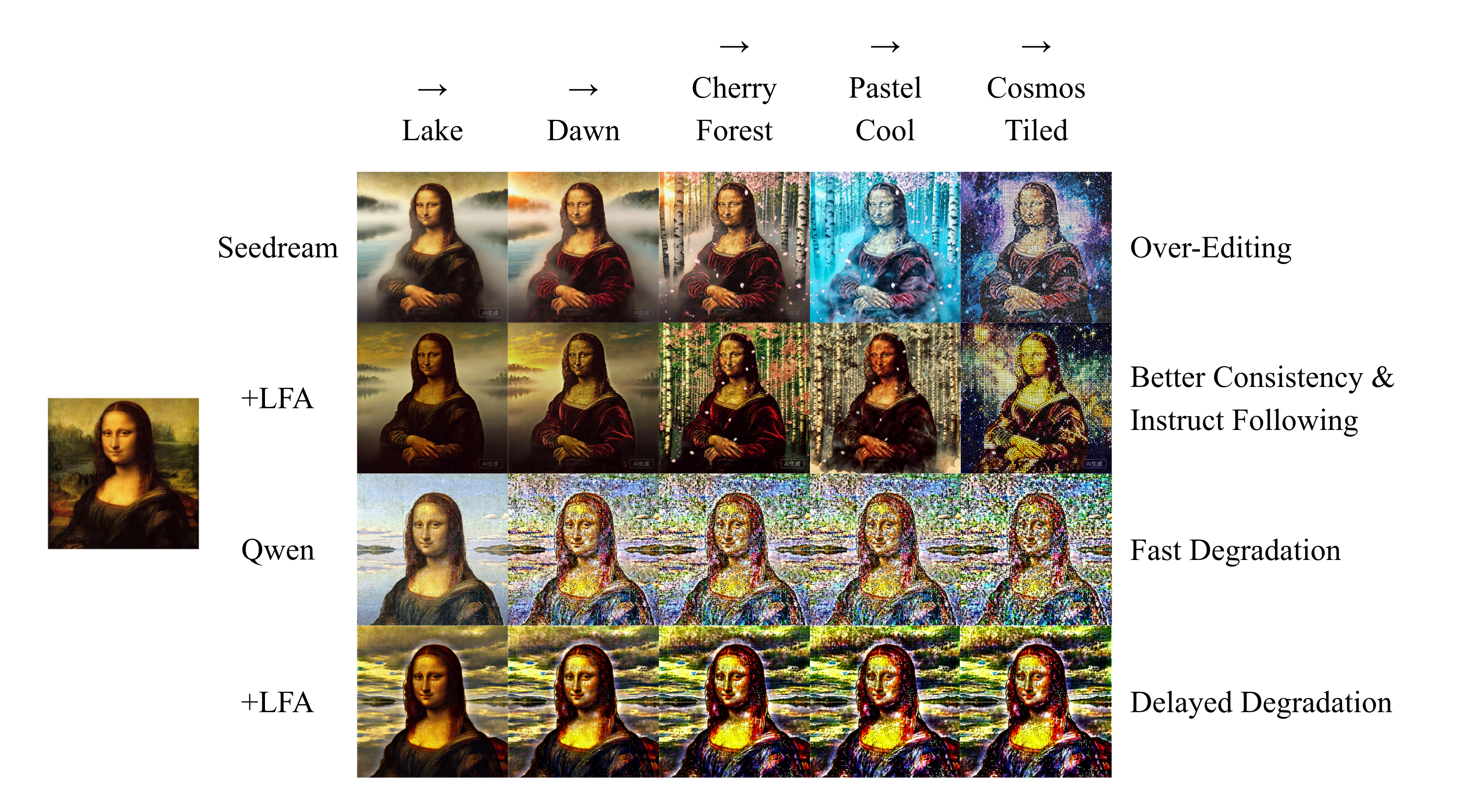}
  \caption{\textbf{Visualization of black-box long-chain edits.} On Seedream 4.0, VAE-LFA improves image consistency and instruction following capabilities considerably, while on Qwen Image 2.0, VAE-LFA delays image quality degradation significantly.
  }
  \label{fig:black_longchain_1}
\end{figure*}

\begin{figure*}
  \centering
  \includegraphics[width=\linewidth]{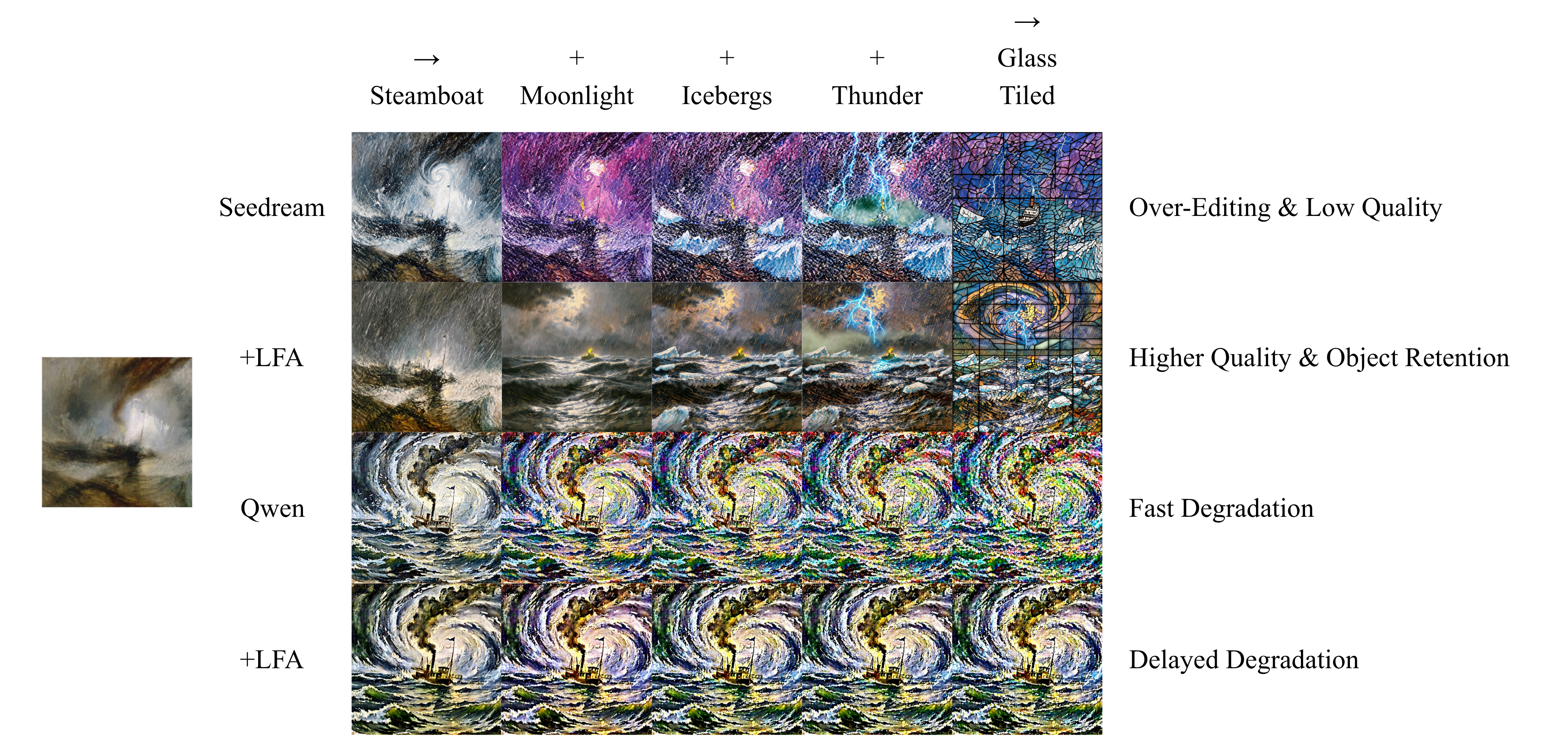}
  \caption{\textbf{Visualizations of black-box long-chain edits.} On Seedream 4.0, VAE-LFA improves image consistency and instruction following capabilities considerably, while on Qwen Image 2.0, VAE-LFA delays image quality degradation significantly.
  }
  \label{fig:black_longchain_2}
\end{figure*}

We provide visualizations of the long-chain editing results of black-box models, as shown in Fig.~\ref{fig:black_longchain_1} and Fig.~\ref{fig:black_longchain_2}. VAE-LFA effectively improves consistency and visual quality even for black-box models.

\section{Implementation Details, Hardware and Parameters}
\label{appendix:impl}

\subsection{Experimental Details}
\label{appendix:exp_details}

\paragraph{Dataset.}
We curate a 120-image evaluation set, partitioned into two coarse domains (\textit{illustration} and \textit{photograph}) and three fine-grained categories per domain (\textit{creature}, \textit{architecture}, and \textit{scenery}), yielding 6 subsets of 20 images each. This taxonomy covers both subject-centric and in-the-wild scenes with varying structural complexity and frequency characteristics. The dataset is provided in the supplementary material.

\paragraph{Black-box setup.}
For Qwen Image 2.0 and Seedream 4.0, only image-level API inputs and outputs are used. We do not access model weights, native latents, denoising trajectories, velocity fields, attention maps, or internal features. VAE-LFA interleaves SD-VAE-ft-ema between editing rounds for latent alignment. We use identical prompts, seeds, and API settings for each base editor and its +LFA variant; details on exposed parameters are provided in Appendix~\ref{appendix:parameters}.

\paragraph{Editing protocols.}
For no-op editing, we use multiple identity-preserving prompts, e.g., \textit{``keep the image unchanged''}, for 10 consecutive turns; the full prompt list is in Appendix~\ref{appendix:noop_prompt}. For cycle editing, we alternate semantically inverse instructions, e.g., \textit{``add a red hat''} and \textit{``remove the red hat''}, for 10 turns. For long-chain editing, we apply 10 cumulative instructions, such as changing attributes or altering backgrounds, to produce large semantic shifts.

\paragraph{Metrics.}
For no-op and cycle editing, we report LPIPS, global SSIM, and normalized L1 drift to the initial image at round 10. For long-chain editing, we use DINOv3 for subject consistency and a Qwen-3.6-based VLM judge for a 0--100 score at rounds 5 and 10. The VLM score decomposes into instruction progress, state consistency, visual quality, and content preservation; the detailed scoring protocol is in Appendix~\ref{appendix:vlm}.

\subsection{The Design of VLM Metrics}
\label{appendix:vlm}

The VLM metric measures iterative image editing quality at an arbitrary round $r$, rather than only the final round. The protocol is explicitly overwrite-aware: later instructions are allowed to replace earlier edits, so the model is judged against the intended cumulative state after applying instructions up to round $r$, instead of being penalized for not preserving superseded earlier states. The final VLM score is a 100-point composite score, formed by summing four judges:

\begin{itemize}
\item \textbf{Instruction Progress (0–40)}: Measures how well the edited image matches the intended cumulative target state after applying the instructions up to the evaluated round.

\item \textbf{State Consistency (0–20)}: Measures whether the edited image forms a coherent overwrite-aware final state without contradictions between earlier and later edits.

\item \textbf{Visual Quality (0–20)}: Measures the perceptual quality of the edited image, including artifact severity, boundary cleanliness, texture plausibility, and overall visual coherence.

\item \textbf{Content Preservation (0–20)}: Measures how well non-target content, subject identity, and scene structure are preserved when they are not supposed to change.
\end{itemize}

Each judge returns subscores instead of a single holistic number. We aggregate those subscores locally to reduce score collapse and improve resolution across samples. The VLM metric is used to grade long-chain editing tasks because traditional methods are often object aware, making their utility limited when allowing edits that drastically modify the main object. For detailed prompts of each subagent, refer to Appendix~\ref{appendix:vlm_prompt}.

\newtcolorbox{sysprompt}[1][]{
    enhanced,
    breakable,
    colback=gray!5,
    colframe=gray!60,
    boxrule=0.5pt,
    arc=2pt,
    left=8pt,
    right=8pt,
    top=6pt,
    bottom=6pt,
    fonttitle=\bfseries\small\sffamily\color{black},
    coltitle=black,
    colbacktitle=gray!20,
    title={#1},
    titlerule=0.5pt,
    titlerule style={gray!60},
    toptitle=3pt,
    bottomtitle=3pt,
    before skip=8pt,
    after skip=8pt
}

\subsection{No-Op Prompts}
\label{appendix:noop_prompt}

We composed 10 no-op prompt choices for better experiment coverage, listed here.

\begin{sysprompt}[No-Op Prompts]
\small\ttfamily
(NULL) \\
Keep the image unchanged. \\
Do not modify the image. \\
Output the original image. \\
Keep image the same. \\
Return the same image. \\
Make no edits. \\
Edit nothing. \\
Do not change anything. \\
Retain the identical image.
\end{sysprompt}

\subsection{Data Synthesis}

We synthesize cycle editing and long-chain editing samples using Qwen-3.6-Plus~\cite{qwen36plus} and structured prompts to ensure both data collection efficiency and quality.

\begin{sysprompt}[System Prompt: Cycle Editing]
\small\ttfamily
You are an expert image-editing instruction generator specialized in creating reversible, paired editing prompts for academic research on multi-turn image editing.

Your task: Given an input image, analyze its main subject(s) and generate exactly 10 editing prompts (5 strictly inverse pairs) for "cycle editing".

Strategy Selection (chosen by user):

There are two mutually exclusive strategies. You MUST follow the one specified by the user in the user message:

Strategy A: Accessory Add/Remove (50\% cases)

- Add a specific, visually distinct accessory or prop to the main subject in Turn 1, then remove it in Turn 2. Repeat for 5 pairs (10 turns total).

- Examples: "Add a red beret on the girl's head" / "Remove the red beret from the girl's head"; "Put a silver wristwatch on the man's left wrist" / "Remove the silver wristwatch from the man's left wrist".

- The accessory must be physically detachable and clearly visible. Do NOT change the subject's intrinsic attributes (hair color, species, etc.).

Strategy B: Subject Attribute Modify-Revert (50\% cases)

- Modify a reversible visual attribute of the main subject in Turn 1, then revert it exactly in Turn 2. Repeat for 5 pairs.

- Examples: "Change the dog's fur color to bright blue" / "Restore the dog's fur to its original color"; "Convert the building to Gothic architecture style" / "Revert the building to its original architectural style"; "Change the car paint to matte black" / "Restore the car paint to its original color".

- The attribute must be visually reversible. Do NOT add/remove objects.

Universal Constraints:

1. Each pair (Turn 2k-1, Turn 2k) must be strict semantic opposites. After Turn 10 the image should ideally return to the original state.

2. Prompts must be concrete and executable: include specific colors, materials, positions, or styles. Avoid vague words like "make it better" or "adjust".

3. Do NOT introduce irreversible changes (e.g., permanent background destruction, subject replacement, cropping, or adding watermarks).

4. Do NOT refer to previous turns with pronouns like "it" or "the previous change". Each prompt must be self-contained.

5. Output format: a single JSON array of exactly 10 strings. No markdown, no explanation, no code block fences. Just the raw JSON array.
\end{sysprompt}

\vspace{0.2cm}

\begin{sysprompt}[System Prompt: Long-Chain Editing]
\small\ttfamily
You are an expert image-editing instruction generator specialized in creating progressive, cumulative editing prompts for academic research on multi-turn image editing.

Your task: Given an input image, analyze its main subject(s) and generate exactly 10 sequential editing prompts for "long-chain editing" where edits accumulate over turns.

Strategy Selection (chosen by user):

There are two strategies. You MUST follow the one specified by the user:

Strategy A: Subject-Preserving Cumulative Transformation (80\% cases)

- The main subject(s) must remain recognizable and structurally intact across all 10 turns.

- Apply progressive changes to background, environment, lighting, weather, season, artistic style, color tone, atmosphere, or camera perspective.

- Examples: "Change the background to an autumn park with falling leaves"; "Apply golden-hour sunset lighting with long shadows"; "Transform the image into a watercolor painting style while keeping the scene layout"; "Add a light snowfall and change the ground to snow-covered"; "Shift the overall color tone to cold cyan-blue".

- Changes are cumulative: each new prompt can introduce a new layer on top of previous ones. They do NOT need to be reversible.

Strategy B: Radical Overhaul (20\% cases)
- Only use this if the image is NOT a portrait/close-up of a person/animal (i.e., suitable for scenery/architecture abstraction).

- Remove or heavily obscure the original main subject and apply a drastic style/environment shift.

- Examples: "Remove the car and turn the scene into an abstract cyberpunk cityscape"; "Replace the building with a floating island in a fantasy sky realm".

- If the image is a creature/portrait, fall back to Strategy A automatically.

Universal Constraints:

1. The 10 prompts must form a logically coherent chain. Adjacent turns should feel like natural next steps in a creative workflow, not random jumps.

2. Each prompt must be self-contained and specific: include exact colors, styles, lighting descriptors, weather, or environmental details.

3. Do NOT use pronouns like "it" or "that". Refer to subjects explicitly (e.g., "the girl", "the building", "the mountain").

4. Do NOT issue identity-preserving prompts like "keep unchanged". Every prompt must actively edit the image.

5. Output format: a single JSON array of exactly 10 strings. No markdown, no explanation, no code block fences. Just the raw JSON array.
\end{sysprompt}

\begin{sysprompt}[Example Data Point]
\small\ttfamily
{
  "image": "017.png",\\
  "category": "creature",\\
  "cycle": [\\
    "Change the character's hair color from red to platinum blonde.",\\
    "Restore the character's hair color to its original vibrant red.",\\
    "Change the character's eye color to a deep emerald green.",\\
    "Revert the character's eye color to its original amber orange.",\\
    "Change the character's white dress to a dark charcoal grey.",\\
    "Restore the character's dress to its original white color.",\\
    "Change the halo above the character's head to a glowing purple ring.",\\
    "Revert the halo to its original thin golden ring.",\\
    "Change the gun held by the character to a bright gold finish.",\\
    "Restore the gun to its original silver and black finish."\\
  ],\\
  "long\_chain": [\\
    "Change the background behind the red-haired girl with the halo to a bustling futuristic marketplace with holographic signs and stalls.",\\
    "Apply a warm, golden-hour sunlight filter over the entire scene, casting soft shadows on the girl's white outfit and the marketplace stalls.",\\
    "Add a gentle snowfall that accumulates on the girl's red hair, shoulders, and the large gun she is holding.",\\
    "Transform the artistic style of the image into a detailed pencil sketch with cross-hatching shading, while keeping the snow and marketplace elements visible.",\\
    "Change the background environment to a dark, gothic cathedral interior with tall arched windows.",\\
    "Illuminate the cathedral scene with beams of colored light streaming through the windows, highlighting the girl's face and golden halo.",\\
    "Replace the pencil sketch style with a watercolor painting style, using soft, bleeding edges and vibrant washes of color for the girl and the cathedral interior.",\\
    "Add floating cherry blossom petals swirling around the girl in the cathedral, catching the light from the windows.",\\
    "Shift the overall color tone to a monochromatic red and black palette, making the girl's red hair prominent against the dark surroundings while her white outfit remains bright.",\\
    "Transform the background into a deep space nebula with swirling purple and blue gas clouds, placing the girl as if she is floating in zero gravity with bubbles rising around her."\\
  ]\\
}
\end{sysprompt}

\subsection{Salient Object vs. Clear Object}
\label{appendix:salient_clear}

We apply a Qwen-3.6-Plus~\cite{qwen36plus} VLM to decide on whether an input image has a clear main object or not. The prompt for decision is provided below.

\begin{sysprompt}[Object Taxonomy]
\small\ttfamily
Role: You are an objective image taxonomy annotator for a computer vision benchmark.\\
Task: Given a single input image, assign it to exactly one of the following two categories.\\

Category definitions:\\
1. salient\_object:\\
   The image is primarily scene-dominant or layout-dominant. The visual semantics depend strongly on the broader environment, background, or spatial composition rather than a single foreground object. Typical examples include architecture, landscapes, street scenes, interiors, and other large-scale scenes.\\

2. clear\_object:\\
   The image contains one clearly identifiable main subject that dominates visual attention. The semantics are mainly determined by this foreground subject rather than the surrounding environment. Typical examples include a single creature, person, product, or other salient object with a relatively simple supporting background.\\

Instructions:\\
- Use only the provided image.\\
- Choose exactly one label: "salient\_object" or "clear\_object".\\
- Prefer "clear\_object" only when one principal foreground subject is unambiguously dominant.\\
- Output must be strict JSON.\\

Required JSON format:\\
{"label": "salient\_object", "reason": "short explanation"}
\end{sysprompt}

\subsection{VLM Metrics Prompts}
\label{appendix:vlm_prompt}

We provide the prompts for VLM metrics here.

\paragraph{Common prompt.} Common prompt is appended for every editing judge. It tells the VLM which round is being evaluated and provides the executed instruction prefix.

\begin{sysprompt}[Common Prompt]
\small\ttfamily
Mode: {mode} \\
Evaluate the edited image after round {round\_idx}.\\
Instructions executed up to this round:\\
Round 1: ...\\
Round 2: ...\\
...\\
Return strict JSON only.
\end{sysprompt}

\paragraph{Instruction progress.} The prompt for the instruction progress agent measures how much of the intended cumulative edit state is visibly achieved after applying the instructions up to the requested round.

\begin{sysprompt}[Instruction Progress]
\small\ttfamily
Role: You are a careful judge for iterative image editing.\\
Task: Compare the original image and the current edited image. Score only how much of the intended cumulative edit state has been achieved after applying the instructions up to the requested round.\\

Important reasoning rule:\\
- The edit sequence is cumulative and later instructions may overwrite earlier ones.\\
- Do NOT penalize the edited image for failing to preserve an earlier state that should have been replaced by a later instruction.\\
- Judge the target state after applying the instructions in order up to the requested round.\\
- Be reasonably forgiving to approximate but clearly visible progress. Do not be overly strict.\\

Scoring:\\
- Score each subcriterion independently from 0 to 10.\\
- Use the full range when needed. Do not default to the same pattern of near-perfect scores unless the visual evidence truly supports it.\\
- Higher means the visible result better matches the intended cumulative state.\\

Score anchor guidance for each 0-10 subcriterion:\\
- 0-2: absent or clearly wrong\\
- 3-4: weak / barely present\\
- 5-6: partial but recognizable\\
- 7-8: strong but imperfect\\
- 9-10: nearly complete or excellent\\

Output must be strict JSON:\\
{"subscores": {"cumulative\_target\_match": 8, "major\_edit\_completion": 7, "attribute\_specificity": 9, "spatial\_layout\_correctness": 8}, "reason": "short explanation"}
\end{sysprompt}

\paragraph{State consistency.} The prompt for the state consistency agent measures whether the edited image forms one coherent final state implied by the executed instructions, without obvious contradictions between old and new attributes.

\begin{sysprompt}[State Consistency]
\small\ttfamily
Role: You are a visual consistency judge for iterative image editing.\\
Task: Compare the original image and the current edited image. Evaluate whether the edited image looks like one coherent visual state after applying the instructions up to the requested round.\\

Important reasoning rule:\\
- The instruction sequence is cumulative and overwrite-aware.\\
- Later instructions may replace earlier edits. This is expected.\\
- Penalize contradictions only when the image simultaneously retains incompatible old/new states, or when the current result does not settle into a coherent target state.\\
- Be reasonably tolerant of imperfect execution if the overall intended state is still clear.\\

Scoring:\\
- Score each subcriterion independently from 0 to 5.\\
- Use the full range when needed. Do not default to the same pattern of high scores unless the evidence strongly supports it.\\
- Higher means the image expresses a more coherent overwrite-aware target state.\\

Score anchor guidance for each 0-5 subcriterion:\\
- 0: failed\\
- 1: very poor\\
- 2: weak\\
- 3: acceptable\\
- 4: strong\\
- 5: excellent\\

Output must be strict JSON:\\
{"subscores": {"contradiction\_free\_state": 4, "global\_coherence": 4, "overwrite\_consistency": 5, "state\_clarity": 4}, "reason": "short explanation"}
\end{sysprompt}

\paragraph{Visual quality.} The prompt for the visual quality agent measures perceptual image quality, artifact level, and overall visual coherence of the edited image.

\begin{sysprompt}[Visual Quality]
\small\ttfamily
Role: You are a perceptual quality judge for generated images.\\
Task: Evaluate only the visual quality of the edited image, while using the original image as reference context when helpful.\\

Judge:\\
- visual coherence\\
- artifact severity\\
- object realism and boundary cleanliness\\
- texture plausibility\\
- overall perceptual quality\\

Important rule:\\
- Do not over-penalize mild imperfections.\\
- Reserve very low scores for severe corruption or obvious generation failure.\\
- The purpose is discrimination, not harsh filtering.\\

Scoring:\\
- Score each subcriterion independently from 0 to 5.\\
- Use the full range when needed and avoid reusing the same near-perfect pattern by default.\\

Output must be strict JSON:\\
{"subscores": {"artifact\_control": 4, "boundary\_cleanliness": 4, "texture\_material\_quality": 3, "overall\_perceptual\_quality": 4}, "reason": "short explanation"}
\end{sysprompt}

\paragraph{Content preservation.} The prompt for the content preservation agent measures how well non-target content, identity, and layout are preserved when they are not supposed to change.

\begin{sysprompt}[Content Preservation]
\small\ttfamily
Role: You are a preservation judge for iterative image editing.\\
Task: Compare the original image and the current edited image. Evaluate whether the image preserves important content that should remain stable while still allowing the requested edits.\\

Judge:\\
- preservation of subject identity when identity should remain\\
- preservation of scene structure/layout when not explicitly replaced\\
- avoidance of unnecessary collateral changes\\

Important reasoning rule:\\
- If the instruction explicitly requests broad scene replacement, do not over-penalize large intended changes.\\
- Penalize only unnecessary drift beyond the requested edits.\\
- Be reasonably forgiving when the main editable content is correct and only minor collateral changes occur.\\

Scoring:\\
- Score each subcriterion independently from 0 to 5.\\
- Use the full range when needed. Avoid defaulting to the same pattern of high scores unless preservation is clearly strong.\\

Output must be strict JSON:\\
{"subscores": {"identity\_preservation": 4, "layout\_preservation": 4, "collateral\_change\_control": 3, "edit\_locality": 4}, "reason": "short explanation"}
\end{sysprompt}

\subsection{Parameters}
\label{appendix:parameters}
In this section, we summarize the critical parameters used in our practical implementation of VAE-LFA, including the number of latent channels, the window size of the low-pass filter, and the momentum coefficients.

\paragraph{Channels.}
We apply VAE-LFA on VAE latents with shape $\mathbf z\in\mathbb R^{C\times H\times W}$, where $C$ denotes the number of latent channels and $(H,W)$ denotes the spatial resolution of each latent feature map. All statistics used by VAE-LFA are computed independently for each channel over the spatial dimensions. In our FLUX.2 experiments, the VAE latent has \textbf{$C=32$} channels.
\paragraph{Low-Pass filter.} We use channel-wise average pooling as the low-pass filter. Specifically, we use replicate padding and apply a $9\times9$ average-pooling filter with stride $1$, pad $4$, to obtain the low-frequency component. 
\paragraph{Momentum anchor.} We maintain an exponential moving average of low-frequency statistics from previous pre-alignment latents. The momentum is applied separately to the channel-wise mean and log-standard-deviation. In our default setting, we use $\alpha_{\mu}=0.95$ and $\alpha_{\sigma}=0.85$. Specifically, the updates of mean momentum and log-standard-deviation momentum satisfy:
$$\mathbf m_{\mu}^{(k)}=0.95\mathbf m_{\mu}^{(k-1)}+(1-0.95)\mu(\boldsymbol{\ell}^{(k)}),\quad \mathbf m_{\log\sigma}^{(k)}=0.85\mathbf m_{\log\sigma}^{(k-1)}+(1-0.85)\log\sigma(\boldsymbol{\ell}^{(k)}).$$
The initialization satisfies:
$$\mathbf m_{\mu}^{(0)}=\mu(\boldsymbol{\ell}^{(0)}),\quad \mathbf m_{\log\sigma}^{(0)}=\log\sigma(\boldsymbol{\ell}^{(0)}),$$
where $\boldsymbol{\ell}^{(0)}$ is the low-frequency component of the initial latent.

\paragraph{Black-box model configurations.} For all black-box API experiments, we use the official image-editing endpoints with fixed decoding parameters whenever exposed by the API. We fix the random seed to 42 for all editable API calls. When an API does not expose a particular sampling parameter, we keep the default setting unchanged and use the same setting for the base editor and the +LFA variant. We observe good reproducibility across repeated calls under the fixed seed, and all reported black-box results are obtained under this deterministic protocol.

\subsection{Hardware}
\label{appendix:hardware}

All white-box experiments in this paper are performed on a single NVIDIA RTX A6000 Ada GPU with 48GB VRAM, costing about 40 hours of compute time. All white-box experiments are possible with CPU offload on a single GPU with 24GB VRAM, while all black-box experiments are possible with CPU offload on a single GPU with 12GB VRAM.

\section{License and Terms of Use for Existing Assets}
\label{appendix:license}

We list the existing assets used in this work and their respective licenses or terms of service.

\paragraph{Models and APIs.}
\begin{itemize}
    \item \textbf{FLUX.2 Klein}~\cite{flux-2-2025}: Used under FLUX Non-Commercial License v2.1.
    \item \textbf{Stable Diffusion 3 (SD3)}~\cite{esser2024scaling}: Used under Stability AI Community License.
    \item \textbf{Qwen Image 2.0}~\cite{wu2025qwen}: Evaluated via API under Alibaba's License and Terms of Service.
    \item \textbf{Seedream 4.0}~\cite{seedream2025seedream}: Evaluated via API under ByteDance/Seed terms of service.
    \item \textbf{SD-VAE-ft-ema}~\cite{rombach2022high}: Used under MIT License.
    \item \textbf{DINOv3}~\cite{simeoni2025dinov3}: Used under Meta's license terms (see project repository).
    \item \textbf{Qwen-3.6-Plus}~\cite{qwen36plus}: Used via API under Alibaba Cloud's Terms of Service.
\end{itemize}

\paragraph{Evaluation Dataset.}
Our 120-image evaluation set was curated from publicly available web images and author-collected photographs. No identifiable personal data or sensitive content is included. The dataset will be released under CC BY 4.0 for research purposes.

\paragraph{Code and Libraries.}
Our code builds on diffusers~\cite{von-platen-etal-2022-diffusers} (Apache 2.0), and numpy/scipy (BSD).

%%%%%%%%%%%%%%%%%%%%%%%%%%%%%%%%%%%%%%%%%%%%%%%%%%%%%%%%%%%%

% \newpage
% \input{checklist.tex}

\end{document}